\documentclass[journal]{IEEEtran}
\usepackage{amsfonts}
\usepackage[nocompress]{cite}
\usepackage[cmex10]{amsmath}
\usepackage{color}
\usepackage{flushend}
\usepackage{url}
\usepackage{array}

\usepackage[table,xcdraw]{xcolor}
\usepackage{lineno}
\usepackage{flushend}
\usepackage{amsfonts}
\usepackage{mdwmath}
\usepackage{booktabs}
\usepackage{rotating}

\definecolor{rblue}{rgb}{0,0.5,1}
\usepackage[colorlinks]{hyperref}
\hypersetup{linkcolor=red,citecolor=rblue}
\usepackage{subfigure}   
\usepackage{graphicx}    
\usepackage{multirow}
\usepackage{makecell}
\usepackage{float}
\usepackage{dsfont}

\begin{document}
\title{AdaptiveClick: Click-aware Transformer with Adaptive Focal Loss for Interactive\\Image Segmentation}

\author{Jiacheng Lin,
        Jiajun Chen,
        Kailun Yang,
        Alina Roitberg,~\IEEEmembership{Member,~IEEE},
        Siyu Li,
        Zhiyong Li,~\IEEEmembership{Member,~IEEE},\\and Shutao Li,~\IEEEmembership{Fellow,~IEEE}
\thanks{This work was supported in part by the National Key Research and Development Program of China under Grant 2022YFB4701404; in part by the National Natural Science Foundation of China under Grant U21A20518 and Grant U23A20341; and in part by Hangzhou SurImage Technology Company Ltd. \emph{(Corresponding authors: Kailun Yang and Zhiyong Li).} }
\thanks{J. Lin and Z. Li are with the College of Computer Science and Electronic Engineering, Hunan University, Changsha 410082, China. 
 (e-mail: jcheng\_lin@hnu.edu.cn; zhiyong.li@hnu.edu.cn).}%
\thanks{J. Chen, S. Li, and K. Yang are with the School of Robotics and the National Engineering Research Center of Robot Visual Perception and Control Technology, Hunan University, Changsha 410082, China. (e-mail: chenjiajun@hnu.edu.cn; lsynn@hnu.edu.cn; kailun.yang@hnu.edu.cn.)}%
\thanks{A. Roitberg is with the Institute for Artificial Intelligence, the University of Stuttgart, Stuttgart 70569, Germany. (e-mail: alina.roitberg@ki.uni-stuttgart.de).}%
\thanks{S. Li is with the College of Electrical and Information Engineering and with the Key Laboratory of Visual Perception and Artificial Intelligence of Hunan Province, Hunan University, Changsha 410082, China. (e-mail: shutao\_li@hnu.edu.cn).}%
}

\markboth{IEEE Transactions on Neural Networks and Learning Systems, March~2024}%
{Lin \MakeLowercase{\textit{et al.}}: AdaptiveClick}

\maketitle

\begin{abstract}
Interactive Image Segmentation (IIS) has emerged as a promising technique for decreasing annotation time. Substantial progress has been made in pre- and post-processing for IIS, but the critical issue of interaction ambiguity, notably hindering segmentation quality, has been under-researched. To address this, we introduce \textsc{AdaptiveClick} -- a click-aware transformer incorporating an adaptive focal loss that tackles annotation inconsistencies with tools for mask- and pixel-level ambiguity resolution. To the best of our knowledge, AdaptiveClick is the first transformer-based, mask-adaptive segmentation framework for IIS. The key ingredient of our method is the Click-Aware Mask-adaptive transformer Decoder (CAMD), which enhances the interaction between click and image features. Additionally, AdaptiveClick enables pixel-adaptive differentiation of hard and easy samples in the decision space, independent of their varying distributions. This is primarily achieved by optimizing a generalized Adaptive Focal Loss (AFL) with a theoretical guarantee, where two adaptive coefficients control the ratio of gradient values for hard and easy pixels. Our analysis reveals that the commonly used Focal and BCE losses can be considered special cases of the proposed AFL. With a plain ViT backbone, extensive experimental results on nine datasets demonstrate the superiority of AdaptiveClick compared to state-of-the-art methods. The source code is publicly available at \url{https://github.com/lab206/AdaptiveClick}.

\end{abstract}


\IEEEpeerreviewmaketitle

\section{Introduction}
\label{sec:introduction}
\IEEEPARstart{I}{nteractive} Image Segmentation (IIS) tasks are designed to efficiently segment a specified object in an image by utilizing a minimal number of user operations, such as drawing boxes~\cite{wu2014milcut, wang2018probabilistic}, scribbling~\cite{wang2020GML, chen2023scribbleseg}, and clicking~\cite{jian2016interactive, sofiiuk2022reviving, liu2022pseudoclick, lin2022focuscut, wang2023one}.
Thanks to the unique interactivity and time-efficiency of this labeling paradigm, existing IIS models are widely or potentially used in fields such as medical image processing~\cite{wang2018deepigeos, diaz2022deepedit}, dataset production~\cite{lin2022focuscut, ding2022deep}, and industry perception~\cite{deng2021application,gu2023precise,li2023human}.

\begin{figure}[t]
\centering

\includegraphics[width=0.95\linewidth]{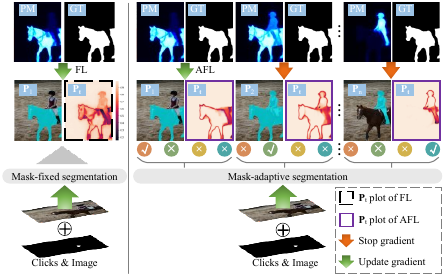}
\caption{Illustration of the proposed method compared with the existing mask-fixed IIS methods. In mask-fixed IIS methods, only a single mask is generated given the input. In contrast, our mask-adaptive AdaptiveClick can produce multiple candidate masks ($\mathbf{P}_\mathrm{1}\!\!\sim\!\!\mathbf{P}_\mathrm{n}$) to address possible ambiguities introduced by user clicks. The model then selects the optimal combination between the Ground Truth (GT) and Probability Map (PM). Finally, Adaptive Focal Loss (AFL) adaptively adjusts the optimal combination to produce a higher-quality mask. Here, $\mathbf{P}_{\mathrm{t}}$ denotes the confidence of the pixel in the sample, with darker colors indicating more hard to segment and vice versa.}

\label{fig:fig1}
\end{figure}

A significant portion of interaction segmentation research targets \textit{click-based} IIS.
Existing studies explored click-based IIS tasks mainly from the perspective of data embedding~\cite{sofiiuk2022reviving, lin2020first, majumder2019SBD}, interaction ambiguity~\cite{chen2021CDNet, li2018LD, liew2019multiseg}, segmentation network~\cite{chen2022focalclick, liu2022isegformer}, post-processing~\cite{chen2022focalclick, lin2022focuscut}, back-propagation~\cite{jang2019BRS, sofiiuk2020f-brs}, and loss optimization~\cite{sofiiuk2022reviving, lin2024CCF}, which have effectively improved training convergence and stability, yielding impressive results.
Although interaction ambiguity has been investigated in earlier studies~\cite{li2018LD, liew2019multiseg, liu2022pseudoclick}, multiple deeper underlying issues, inter-class click ambiguity and intra-class click ambiguity, have hindered effective solutions.

On the one hand, inter-class click ambiguity arises when a click may correspond to multiple potential objects in an image. For instance, in Fig.~\ref{fig:fig1}, the potential object could be a \emph{human}, a \emph{horse}, or a \emph{combination of both} with a click. However, most conventional IIS methods are mask-fixed methods, which can only produce a single mask, rendering them ineffective in addressing the click ambiguity. Apart from the ambiguity introduced by user clicks, another significant factor exacerbating inter-class click ambiguity is the long-range propagation fading of click features. On the other hand, intra-class click ambiguity is induced by ``gradient swamping'' during the optimization process of Focal Loss (FL). This results in a substantial number of misclassified pixels around the decision boundary, as illustrated by the $\mathbf{P}_{\mathrm{t}}$ plot of FL in Fig.~\ref{fig:fig1}, further exacerbating the interaction ambiguity. The ``gradient swamping'' refers to the phenomenon of FL focusing too much on the classification of hard pixels, significantly weakening the gradient values that many low-confidence easy pixels (also referred to as ambiguous pixels) should contribute.

In this paper, we rethink the interaction ambiguity of IIS by addressing both inter-class and intra-class click ambiguities. Firstly, we propose a mask-adaptive AdaptiveClick, which is composed of a Click-Aware Mask-adaptive transformer Decoder (CAMD) with an Adaptive Focal Loss (AFL).
CAMD incorporates a click-aware attention module that generates distinct instance masks for each click by considering potential click ambiguities and subsequently selecting the optimal mask. This addresses inter-class click ambiguity and accelerates the convergence of the transformer. 
Next, we observe that the root of intra-class click ambiguity is ``gradient swamping'', which aggravates interaction ambiguity in IIS tasks. 
To tackle this, we put forward a novel AFL based on the gradient theory of BCE and FL. 
AFL adapts the training strategy according to the difficulty distribution of samples, improving intra-class click ambiguity problems (the $\mathbf{P}_{\mathrm{t}}$ plot of AFL in Fig.~\ref{fig:fig1} with clearer boundaries).

At a glance, the main contributions delivered in this work are summarized as follows:

1) A novel mask-adaptive segmentation framework is designed for IIS tasks. To the best of our knowledge, this is the first mask-adaptive segmentation framework based on transformers in the context of IIS;

2) We put forward a new clicks-aware mask-adaptive transformer decoder with a clicks-aware attention module to tackle the problem of inter-class click ambiguity arising from user ambiguity clicks and the long-range propagation fading of clicks features in the IIS methods;

3) A novel adaptive focal loss is designed to overcome the ``gradient swamping'' problem specific to focal loss-based training in IIS, which accelerates model convergence and alleviates intra-class click ambiguity;

4) Experimental results from nine datasets showcase the clear benefits of AdaptiveClick and adaptive focal loss, yielding state-of-the-art performance on IIS tasks.

The remainder of the paper is structured as follows: Sec.~\ref{sec:related work} provides a brief overview of related work. The proposed methods are presented in Sec.~\ref{sec:method}. We present experimental results in Sec.~\ref{sec:experiments}, followed by the discussion in Sec.~\ref{sec:discussion}. Sec.~\ref{sec:conclusion} summarizes the findings presented in this work.

\begin{figure*}[t]
\centering
\includegraphics[width=0.9\linewidth]{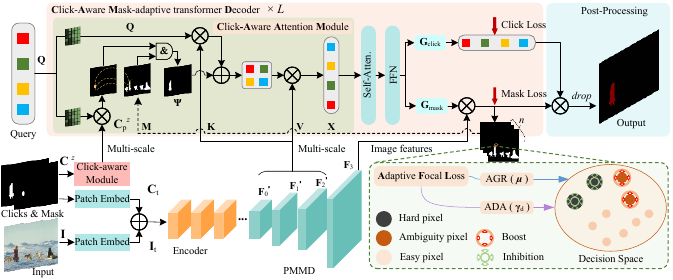}	
\caption{An illustration of AdaptiveClick. First, clicks with the previous mask and image ($\mathbf{I}$) features are obtained via patch embedding, then fused by addition. Second, the fusion features are obtained by Encoder, and the pixel features $\mathbf{F}_0$, $\mathbf{F}_1$, and $\mathbf{F}_2$ of different dimensions are obtained by Pixel-level Multi-scale Mask transformer Decoder (PMMD). Third, $\mathbf{F}^{'}_0, \mathbf{F}^{'}_1, \mathbf{F}^{'}_2$, and $\mathbf{F}_3$ and clicks are jointly input into the designed Click-Aware Mask-adaptive transformer Decoder (CAMD). Fourth, CAMD generates $n$ corresponding masks $\hat{p}$ based on each click and then completes the optimization training process with ADA and AGR proposed in Adaptive Focal Loss (AFL). Finally, the obtained mask sequences are passed through post-processing to output the final mask.}
\label{fig:fig2}
\end{figure*}

\section{Related Work} \label{sec:related work}
\subsection{Architecture of Interactive Image Segmentation}

In the IIS task, the goal is to obtain refined masks by effectively leveraging robust fusion features activated by click information.
Interaction strategies in mainstream research works can be generally divided into three categories: pre-fusion~\cite{xu2016DIOS,lin2020first,sofiiuk2022reviving,liu2022simpleclick}, secondary fusion~\cite{ding2022deep,lin2020first,chen2021CDNet,wei2023focused} and middle-fusion~\cite{kirillov2023segany,huang2023interformer,RanaMahadevan23DynaMITe,zou2023segment}.

Pre-fusion incorporates click as an auxiliary input, which includes click features~\cite{xu2016DIOS} and previous mask~\cite{sofiiuk2022reviving,lin2022focuscut,liu2022simpleclick}.
Starting with click embedding via distance transformation~\cite{xu2016DIOS}, a common technique is to concatenate maps of clicks with raw image~\cite{jang2019BRS,li2018LD,liew2017regional}.
Subsequently, Benenson~\textit{et al.}~\cite{benenson2019large} suggest that using clicks with an appropriate radius offers better performance. Nevertheless, due to the potential absence of pure interaction strategies in~\cite{xu2016DIOS}, another line of work~\cite{forte2020getting,sofiiuk2022reviving} constructively employs masks predicted from previous clicks as input for subsequent processing. 
Following this simple yet effective strategy, several studies have achieved remarkable results, such as incorporating cropped focus views~\cite{lin2017focal} or exploiting IIS transformer~\cite{liu2022simpleclick}.
Although the above methods strive to maximize the use of click, the click feature tends to fade during long-range propagation, exacerbating the interaction ambiguity of IIS methods.

To address the above issues, secondary fusion has been introduced as a potential solution. 
Lin~\textit{et al.}~\cite{lin2020first} emphasizing the impact of the first click, long-range propagation strategy for click~\cite{chen2021CDNet}, and utilizing multi-scale strategies~\cite{liew2019multiseg}.
Inspired by~\cite{sofiiuk2022reviving}, Chen~\textit{et al.}~\cite{chen2022focalclick} further fuse click and previous mask for the refinement of the local mask.
Recently, Wei~\textit{et al.}~\cite{wei2023focused} proposed a deep feedback-integrated strategy that fuses coordinate features with high-level features.
Moreover, recent works~\cite{jang2019BRS,sofiiuk2020f-brs,kontogianni2020IA-SA} have combined click-embedding strategies~\cite{xu2016DIOS} with an online optimization approach.
However, no existing research has explored overcoming the challenge of interaction ambiguity in vision transformers.
 
Recently, middle-fusion strategies have gained popularity in the IIS field.
Kirillov~\textit{et al.}~\cite{kirillov2023segany} propose a novel strategy that treats click as prompt, supporting multiple inputs.
Similarly, Zou~\textit{et al.}~\cite{zou2023segment} concurrently suggest joint prompts for increased versatility.
For better efficiency, Huang~\textit{et al.}~\cite{huang2023interformer} bypass the strategies in~\cite{sofiiuk2022reviving} related to the backbone to perform inference multiple times. Inspired by Mask2Former~\cite{cheng2022masked}, a parallel study~\cite{RanaMahadevan23DynaMITe} models click as queries with the timestamp to support multi-instance IIS tasks. Although these models achieve remarkable success, they sacrifice the performance of a single IIS task to enhance multi-task generalization.

In this paper, we first explore mask-adaptive transformers for IIS, focusing on solving interaction ambiguity.
Specifically, AdaptiveClick generates multiple candidate masks from ambiguity clicks and selects the optimal one for the final inference.
Further, AdaptiveClick achieves long-range click propagation within transformers through a click-aware attention module, breaking the limitations of mask-fixed IIS methods.

\subsection{Loss Function of Interactive Image Segmentation}

The BCE loss~\cite{yi2004automated} and its variants~\cite{lin2017focal,leng2022polyloss,pihur2007weighted,xie2015holistically,li2022equalized} are widely used in IIS tasks~\cite{jang2019BRS,liew2019multiseg,maninis2018deep,lin2024CCF} for training.
However, BCE treats all pixels equally, resulting in a gradient of easy pixels inundated by hard pixels and blocking the model's performance.
Previous efforts~\cite{lin2017focal,song2023dynamic,pihur2007weighted} attempt to solve this issue from the imbalance of positive/negative samples~\cite{rahman2016optimizing,milletari2016v} or easy/hard samples~\cite{lin2017focal,leng2022polyloss}.

To counteract the imbalance between positive and negative samples, WBCE~\cite{pihur2007weighted} introduces a weighting coefficient for positive samples.
Furthermore, Balanced CE~\cite{xie2015holistically} weights not only positive samples but also negative samples.
These methods are used with data that satisfies a skewed distribution~\cite{jadon2020survey} but are blocked in balanced datasets due to their adjustable parameters, which influence the model's performance. 
To address the imbalance between hard and easy samples, the FL~\cite{lin2017focal} formulates a difficulty modifier to enhance model training and down-weight the impact of easy examples, thereby allowing the model to focus more on the hard samples.
Recently, Leng~\textit{et al.}~\cite{leng2022polyloss} offer a new perspective and propose the Poly Loss (PL) as a linear combination of polynomial functions.
Sofiiuk~\textit{et al.}~\cite{sofiiuk2019nfocal} present the Normalized Focal Loss (NFL), which expands an extra correction factor negatively correlated with the total modulate factor in FL.
In addition, some other works~\cite{rahman2016optimizing,milletari2016v} also attempt to solve this problem.
However, deeper reasons, such as the case that many low-confidence pixels are caused by gradient swamping, make ambiguous pixels impossible to be effectively classified.

In brief, it is crucial for the IIS tasks to address the issue of ``gradient swamping'' for FL-based losses to alleviate the interaction ambiguity. Unlike previous works~\cite{jang2019BRS,sofiiuk2020f-brs}, we consider the interaction ambiguity from the perspective of intra-class click ambiguity optimization.
Therefore, our model can focus more on ambiguous pixels for the IIS task than simply computing gradient values point-by-point equally during the gradient backpropagation process.

\section{Method} \label{sec:method}

At a high level, AdaptiveClick's primary components consist of four key stages: (1) data embedding, (2) feature encoding, (3) feature decoding, and (4) loss optimization. These stages are illustrated and summarized in Fig.~\ref{fig:fig2}. In our approach, we address the interaction ambiguity by focusing on aspects of feature decoding and loss optimization.

\subsection{Deficiency of Existing IIS Methods} \label{sec:Pre}

In this paper, we categorize interaction ambiguity into inter-class and intra-class click ambiguity and tackle them by focusing on two distinct aspects: inter-class click ambiguity resolution and intra-class click ambiguity optimization.

\textbf{Inter-class Click Ambiguity Resolution.} Some IIS methods with fixed masks have explored interaction ambiguity by incorporating multi-scale transformations~\cite{li2018LD, liew2019multiseg, liu2022pseudoclick} and investigating long-range propagation of clicks~\cite{chen2021CDNet, lin2020first, ding2022rethinking}. However, these studies primarily address interaction ambiguity by emphasizing click enhancement. While these methods effectively enhance the guiding role of clicks in the segmentation process, they still lack effective tools for handling click ambiguity. Thus, their primary focus is on maximizing the model's ability to generate a single ground-truth-compliant mask by emphasizing the click, but they fall short of effectively addressing click ambiguity.

\textbf{Intra-class Click Ambiguity Optimization.} 
Given the final prediction ($\mathbf{P}_{\mathrm{n}}$) and ground truth ($\mathbf{y}_{\mathrm{t}}$), BCE~\cite{yi2004automated} widely applied in IIS~\cite{sofiiuk2022reviving, jang2019BRS, lin2022focuscut} can be expressed as,
\begin{equation}\label{L_nbce}
\begin{aligned}
\mathbb{\ell }_{\mathrm{bce}} = -\sum_{i=1}^{\mathcal{N} } \mathrm{log}\left(\mathbf{P}^{i}_{\mathrm{t}}\right).
\end{aligned}
\end{equation}
where $\mathbf{P}^{i}_{\mathrm{t}} \!\! = \!\! \left\{\begin{matrix} \mathbf{P}^{i}_{\mathrm{n}}, \mathrm{if} \indent \mathbf{y}^{i}_{\mathrm{t}}=1 \\1-\mathbf{P}^{i}_{\mathrm{n}}, \mathrm{else} \indent \mathbf{y}^{i}_{\mathrm{t}} = 0 \end{matrix}\right.$. $\mathbf{P}^{i}_{\mathrm{t}}\!\! \in \!\! [0,1]$ is difficulty confidence of each pixel and $\mathbf{P}^{i}_{\mathrm{n}} \!\! \in \!\! \mathbf{P}_{\mathrm{n}}$, $\mathbf{P}^{i}_{\mathrm{t}} \!\! \in \!\! \mathbf{P}_{\mathrm{t}}$ and $\mathbf{y}^{i}_{\mathrm{t}} \!\! \in \!\! \mathbf{y}_{\mathrm{t}}$. $\mathcal{N} =HW$, where $H$ and $W$ are the height and width of $\mathbf{P}_{\mathrm{n}}$.

\begin{figure}[tbp]
\centering
\includegraphics[width=0.9\linewidth]{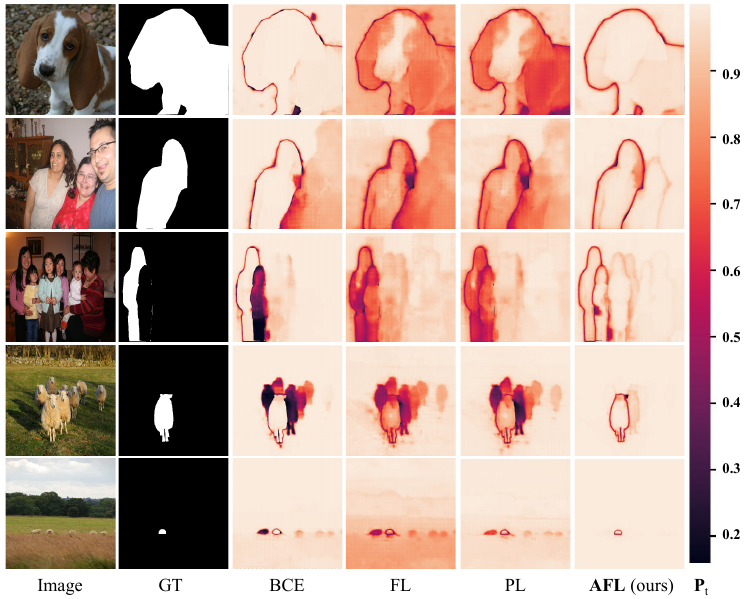}	
\caption{Difficulty confidence visualization of different loss functions on the SBD~\cite{majumder2019SBD} training dataset. From left to right are the image, the ground truth, and the $\mathbf{P}_{\mathrm{t}}$ plot of BCE~\cite{yi2004automated}, FL~\cite{lin2017focal}, PL~\cite{leng2022polyloss}, and  AFL, respectively.}
\label{fig:fig3}
\end{figure}

However, $\mathbb{\ell }_{\mathrm{bce}}$ treats hard and easy pixels equally~\cite{lin2017focal}, we call this property of BCE ``difficulty-equal'' (a theoretical proof in Eq. (\ref{L_dbce})), which will result in many hard pixels not being effectively segmented. To address those issues, FL~($\mathbb{\ell }_{\mathrm{fl}}$)~\cite{lin2017focal} is introduced in IIS works~\cite{liu2022simpleclick,zhou2023interactive,huang2023interformer,RanaMahadevan23DynaMITe}, where,
\begin{equation}\label{L_bce}
\begin{aligned}
\mathbb{\ell }_{\mathrm{fl}} = -\sum_{i=1}^{\mathcal{N} } \left ( 1-\mathbf{P}^{i}_{\mathrm{t}} \right )^{\gamma}\mathrm{log} \left ( \mathbf{P}^{i}_{\mathrm{t}} \right ).
\end{aligned}
\end{equation}
where $(1-\mathbf{P}^{i}_{\mathrm{t}})^{\gamma}$ is a difficulty modifier, $\gamma \!\!\in\!\! [0,5]$, the larger of $\gamma$, the greater weight for hard pixels. $\mathbb{\ell }_{\mathrm{fl}}$ can solve the imbalance of hard and easy pixels due to its ``difficulty-oriented'' property (a theoretical proof in Eq.~(\ref{L_fl_gr})).

However, study~\cite{leng2022polyloss} points out that the coefficients of each Taylor term in Eq.~(\ref{L_bce}) are not optimal, so the PL ($\mathbb{\ell }_{\mathrm{pl}}$) is proposed as a correction scheme to improve the performance of $\mathbb{\ell }_{\mathrm{fl}}$. The $\mathbb{\ell }_{\mathrm{pl}}$ can be expressed as
\begin{equation}\label{L_pl}
\mathbb{\ell }_{\mathrm{pl}}=- \sum_{i=1}^{\mathcal{N} } \left[\left ( 1-\mathbf{P}^{i}_{\mathrm{t}} \right )^{\gamma}\mathrm{log}\left (\mathbf{P}^{i}_{\mathrm{t}} \right ) + \alpha \left ( 1-\mathbf{P}^{i}_{\mathrm{t}} \right )^{\gamma+1} \right].
\end{equation}
where the $\alpha$ is a coefficient of $\mathbb{\ell }_{\mathrm{pl}}$.

Such improvements are beneficial, but as shown in Fig.~\ref{fig:fig3}, we observe that pixels around the click target boundaries (\textit{e.g.}, the person next to the lady in white in the second row) are still difficult to classify correctly.
We define such pixels as low-confidence easy pixels or ambiguous pixels in the IIS task. This phenomenon occurs because both BCE~\cite{yi2004automated} and FL~\cite{lin2017focal} suffer from gradient swamping.
The difference is that the gradient swamping of BCE is mainly the result of a large number of easy pixels swamping the gradients of hard pixels. In contrast, the gradient swamping of FL results from the gradients of ambiguous pixels being overwhelmed by those of hard pixels. This dilemma hinders the excellent performance of ambiguous pixels and further exacerbates the interaction ambiguity of the IIS model. We term this phenomenon ``gradient swamping by FL-based loss of ambiguous pixels in IIS''.

\subsection{Inter-class Click Ambiguity Resolution} \label{sec:afl}

Previous studies~\cite{sofiiuk2022reviving,liu2022simpleclick} have experimented with using prior clicks as prior knowledge to help models converge faster and focus more precisely on local features. Motivated by this, we rethink the  inter-class click ambiguity of mask-fixed based IIS methods and address them by exploring a click-aware mask-adaptive transformer decoder.

\subsubsection{Pixel-level Multi-scale Mask Transformer Decoder} \label{sec:PMTD}

To obtain representations with more subtle differences of different clicks, we expand the fusion features $\mathbf{F}$ to $1/8$, $1/16$, and $1/32$ resolution based on the feature pyramid networks~\cite{lin2017feature}, denoted as $\mathbf{F}_0$, $\mathbf{F}_1$, and $\mathbf{F}_2$. Then, $\mathbf{F}_0$, $\mathbf{F}_1$, and $\mathbf{F}_2$ are passed to a $6$ multi-scale deformable attention transformer~\cite{zhu2021dd}, to obtain $\mathbf{F}^{'}_0$, $\mathbf{F}^{'}_1$, and $\mathbf{F}^{'}_2$, respectively. Finally, a lateral up-sampling layer is used on the $1/8$ feature map to generate $\mathbf{F}_{3}$ with a resolution of $1/4$ as a pixel-by-pixel embedding.

\subsubsection{Click-Aware Mask-adaptive Transformer Decoder} \label{sec:CAMD}

The proposed Clicks-Aware Mask-adaptive transformer Decoder (CAMD) primarily consists of the Click-Aware Attention Module (CAAM), self-attention module, feed-forward network, and the click- and mask prediction heads. 

\textbf{Click-Aware Attention Module.} Given a set feature of positive clicks $\mathbf{C}^{'}_{\mathrm{p}} \in \left \{\mathbf{C}^1_\mathrm{p}, \mathbf{C}^2_\mathrm{p}, ..., \mathbf{C}^z_\mathrm{p} \right \}$ and their corresponding different scales decoding features $\left \{\mathbf{F}^{'}_0, \mathbf{F}^{'}_1, \mathbf{F}^{'}_2\right \}$, where $z$ is the number of positive clicks among total $m$ clicks and $z< m$. Then, the attention matrix of the proposed CAAM can be expressed as,
\begin{equation}\label{click_atten}
\begin{aligned}
\mathbf{X}_{l} = \text{softmax}\left ( \Psi_{l-1}+\mathbf{Q}_{l}\mathbf{K}^{\mathsf{T}}_{l}\mathbf{V}_{l} \right )+\mathbf{X}_{l-1},
\end{aligned}
\end{equation}
where $l$ is the layer index, $\mathbf{X}_l \in \mathbb{R}^{n \times d}$ is the $n \times d$-dimensional query feature of the $l$-th layer, and $\mathbf{Q}_{l} = f_{Q} \left (\mathbf{X}_{l-1} \right ) \in \mathbb{R}^{n \times d}$. $\mathbf{X}_{0}$ denotes the input query feature of the transformer decoder. $\mathbf{K}_l$, $\mathbf{V}_l \in \mathbb{R}^{H_l W_l \times d}$ are the image features under the transformations $f_k$ and $f_v (*)$ from $\left \{\mathbf{F}^{'}_0, \mathbf{F}^{'}_1, \mathbf{F}^{'}_2\right \}$, respectively, and $H _l$ and $W_l$ are the spatial resolutions of the $\left \{\mathbf{F}^{'}_0, \mathbf{F}^{'}_1, \mathbf{F}^{'}_2\right \}$. 
$f_Q$, $f_k$, and $f_v$ are linear transformations. 
$\Psi_{l-1}$ is click attention matrix and can be obtained via Eq.~(\ref{M_atten})
\begin{equation}\label{M_atten}
\begin{aligned}
\Psi_{l-1} = \psi\left [\omega _{f} \left ( \mathbf{C}^{'}_{\mathrm{p}} \right ) \left [\mathbf{Q}_{l} \right ]_{+}^{\mathsf{T}} \right] \& \mathbf{M}_{l-1},
\end{aligned}
\end{equation}
where $\omega _{f}$ is the click-aware module, which consists of a max pooling layer and a linear layer. $\psi$ denotes the linear mapping layer, $[*]_{+}$ represents the max function, $\&$ represents the $and$ operation. $\Psi_0$ is the binary click prediction obtained from $X_0$, 
\textit{i.e.}, before the query features are fed to the transformer decoder. 
$\mathbf{M}_{l-1}$ represents the attention mask of the previous transformer block, which can be calculated by Eq.~(\ref{M_metric})
\begin{equation}\label{M_metric} 
\begin{aligned}
\mathbf{M}_{l-1} \left (i,j \right ) = \left\{\begin{matrix}
0, \indent \mathrm{if} \indent \mathbf{F}_{3} \left (\hat{\mathbf{y}}_{\mathrm{t}}^{\small l-1} \right )^{\mathsf{T}} \left(i,j \right ) = 1	\\
- \infty, \indent \mathrm{otherwise}
\end{matrix}\right. .
\end{aligned}
\end{equation}
where $\mathbf{F}_{3}(\hat{\mathbf{y}}_{\mathrm{t}}^{\small l-1})^{\mathsf{T}}\!\! \in [0,1]^{n \times H_l W_l}$ is the binarized mask with a threshold of $0.5$ resized by the ($l\!\!-\!\!1$)-th transformer decoder.

\textbf{Mask Prediction Head.} The mask prediction head ($\mathbf{G}_{\mathrm{mask}}$) consists of three MLP layers of shape $n\!\! \times \!\!d$. $\mathbf{G}_{\mathrm{mask}}$ takes predicted object query $\mathbf{Q}_{l}$ as input and outputs a set of click-based prediction ($\hat{\mathbf{y}}_{\mathrm{t}}$), where $\hat{\mathbf{y}}_{\mathrm{t}}=\mathbf{G}_{\mathrm{mask}}(\mathbf{Q}_{l})$.

\textbf{Clicks Prediction Head.} The click prediction head ($\mathbf{G}_{\mathrm{click}}$) consists of a linear layer of shape $n \times 2$ with a softmax. Given a $\mathbf{Q}_{l} \in \mathbb{R}^{d}$, $\mathbf{G}_{\mathrm{click}}$ takes $\mathbf{Q}_{l}$ as input and outputs a click prediction ($\hat{c}$), where $\hat{c}=\mathbf{G}_{\mathrm{click}}(\mathbf{Q}_{l})$.

\subsubsection{Mask-adaptive Matching Strategy}

To discriminate and generate a mask without ambiguity, we use the Hungarian algorithm~\cite{cheng2022masked, cheng2021per} to find the optimal permutation $\hat{\mathbb{\sigma}}$ generated between the $\mathbf{P}_{\mathrm{n}}$ and $\mathbf{y}_{\mathrm{t}}$ during the mask-adaptive optimization process, and finally to optimize the object target-specific losses. Accordingly, we search for a permutation $\hat{\mathbb{\sigma}} \in \mathbf{S}_{n_{q}}$ with the lowest total cost,
\begin{equation}\label{L_Match}
\begin{aligned}
\hat{\mathbb{\sigma}} = \underset{\sigma \in \mathbf{S}_{n_{q}}}{\mathrm{arg min}} \sum_{i=1}^{n_{q}} \mathds{1}_{\left \{\hat{\mathbf{y}}_{\mathrm{p}} \ne \emptyset  \right \}} \mathbb{L }_{\mathrm{total}} \left(\hat{\mathbf{y}}_{\mathrm{p}} \mathbb{\sigma}(i), \mathbf{y}^{i}_{\mathrm{p}} \right ).
\end{aligned}
\end{equation}
where $\mathbb{L }_{\mathrm{total}}$ is total loss can be obtained in Eq.~(\ref{L_total}), $\hat{\mathbf{y}}_{\mathrm{p}}$ is the predicted via click instance and $\hat{\mathbf{y}}_{\mathrm{p}}= \left \{\hat{\mathbf{y}}_{\mathrm{t}}^{i} \right \}_{i=1}^{n_{q}}=\left(\hat{\mathbf{y}}_{\mathrm{t}}, c_{m} \right)$. $n_{q}$ is the number of predicted via the click instance and the $m$ is the number of click. Similarly, $\mathbf{y}_{\mathrm{p}}$ is the $\mathbf{y}_{\mathrm{t}}$ via click instance, and $\mathbf{y}_{\mathrm{p}}= \left \{\mathbf{y}_{\mathrm{t}}^{i} \right \}_{i=1}^{n_{q}} = \left(\mathbf{y}_{\mathrm{t}}, c_{m} \right)$.

Inspired by~\cite{cheng2022masked, cheng2021per}, CAMD uses a multi-scale strategy to exploit high-resolution features. It feeds continuous feature mappings from the PMMD to the continuous CAMD in a cyclic manner. It is worth mentioning that the CAMD enables clicks to propagate over a long range and accelerates the model's convergence. This is because, in previous works~\cite{cheng2022masked, cheng2021per}, the optimization of the query had a random property. However, CAMD can correctly pass the click to the fine mask in $\hat{\mathbf{y}}_{\mathrm{p}}$, thus achieving the convergence of the model faster and locating the objects effectively.

\subsection{Intra-class Click Ambiguity Optimization} \label{sec:ambiguity}

There are no studies focused on solving the ``gradient swamping‘’ in image segmentation. Existing methods only explored the imbalance of positive and negative pixels~\cite{pihur2007weighted, xie2015holistically}, and the imbalance of hard and easy pixels~\cite{lin2017focal, sofiiuk2019nfocal, leng2022polyloss} of BCE. Unlike these works, we rethink the gradient swamping of FL and propose Adaptive Focal Loss (AFL) based on the gradient theory of BCE and FL. 
It enables the model to adapt the learning strategy according to the global training situation, mitigating the gradient swamping.

\subsubsection{Adaptive Difficulty Adjustment} \label{sec:sec3.3}

Observing Eq.~(\ref{L_bce}) and Eq.~(\ref{L_pl}), 
$\mathbb{\ell }_{\mathrm{fl}}$ and $\mathbb{\ell }_{\mathrm{pl}}$ use $(1-\mathbf{P}^{i}_{\mathrm{t}})^{\gamma}$ as the difficulty modifier, and adjust the value of $(1-\mathbf{P}^{i}_{\mathrm{t}})^{\gamma}$ using initial $\gamma$ according to the difficulty distribution of the dataset empirically (\textit{i.e.}, generally taken as $2$ in previous works).
However, it is worth noting that the distribution of difficulty varies among samples even in the same dataset (see Fig.~\ref{fig:fig3}).
Therefore, giving a fixed $\gamma$ to all training samples of a dataset is a suboptimal option.

To give the model the ability to adaptively adjust $(1-\mathbf{P}^{i}_{\mathrm{t}})^{\gamma}$ according to the different difficulty distributions of the samples and learning levels, we introduce an Adaptive Difficulty Adjustment (ADA) factor $\gamma_{a}$.
With $\gamma_{a}$, the $(1-\mathbf{P}^{i}_{\mathrm{t}})^{\gamma}$ of each pixel can be adjusted according to the overall training of the samples, giving the model the capability of ``global insight''.

Given $\mathbf{I}$, we refer to the foreground as the hard pixels and the number as $\mathcal{H}$ (obtained through $\mathbf{y}_{\mathrm{t}}$).
To estimate the overall learning difficulty of the model, we use the $\mathbf{P}^{i}_{\mathrm{t}}$ of each pixel to represent its learning level. Then, the learning state of the model can be denoted as $\sum_{i=1}^{\mathcal{H}} (\mathbf{P}^{i}_{\mathrm{t}})$.
Ideally, the optimization goal of the model is to classify all hard pixels completely and accurately. In this case, the optimal learning situation of the model can be represented as $\sum_{i=1}^{\mathcal{H}} (\mathbf{y}^{i}_{\mathrm{t}})$. Given the FL properties and the actual learning situation of the model,  $\mathbb{\gamma}_{\mathrm{a}}$ can be represented as
\begin{equation}\label{L_adm}
\small
\begin{aligned}
\mathbb{\gamma}_{\mathrm{a}} &= 1-\frac{\sum_{i=1}^{\mathcal{H}} (\mathbf{P}^{i}_{\mathrm{t}})}{\sum_{i=1}^{\mathcal{H}} (\mathbf{y}^{i}_{\mathrm{t}})},
\end{aligned}
\end{equation}

Then, the difficulty modifier for AFL is,
\begin{equation}\label{L_rdm}
\begin{aligned}
(1-P_{t})^{\gamma_{\mathrm{d}}}, \text{where} \indent \gamma_{\mathrm{d}} = \gamma+\gamma_{\mathrm{a}},
\end{aligned}
\end{equation}

In summary, AFL can be expressed as:
\begin{equation}\label{L_bfl_v}
\begin{aligned}
\mathbb{\ell }_{\mathrm{afl}} &= \sum_{i=1}^{\mathcal{N} } -(1-\mathbf{P}^{i}_{\mathrm{t}})^{\gamma_{\mathrm{d}}} \mathrm{log}(\mathbf{P}^{i}_{\mathrm{t}}) + \alpha (1-\mathbf{P}^{i}_{\mathrm{t}})^{\gamma_{\mathrm{d}}+1}.
\end{aligned}
\end{equation}

\subsubsection{Adaptive Gradient Representation}

For the proposed AFL, a larger $\gamma_{\mathrm{d}}$ is suitable when a more severe hard-easy imbalance is present. However, for the same $\mathbf{P}_{\mathrm{t}}$, the larger of $\gamma_{\mathrm{d}}$, the smaller the loss. It leads to the fact that when one wants to increase the concentration on learning with a severe hard-easy imbalance, it tends to sacrifice a portion of the low-confidence easy pixel's loss in the overall training process.

To address this issue, we first explore the gradient composition of $\mathbb{\ell }_{\mathrm{afl}}$ and its properties based on Taylor approximation theory~\cite{rudin1976principles} of Eq.~(\ref{L_nbce}), we can obtain Eq.~(\ref{L_fl_ttt}),
\begin{equation}\label{L_bce_ttt}
\begin{aligned}
\mathbb{\ell }_{\mathrm{bce}} &= \sum_{i=1}^{\mathcal{N} } \left [ \left(1-\mathbf{P}^{i}_{\mathrm{t}} \right)+\frac{1}{2} \left(1-\mathbf{P}^{i}_{\mathrm{t}} \right)^{2}+... \right ] ,
\end{aligned}
\end{equation}

To further explore the theoretical connection between the $\mathbb{\ell }_{\mathrm{bce}}$ and proposed $\mathbb{\ell }_{\mathrm{afl}}$, we conducted an in-depth study at the gradient level of Eq.~(\ref{L_bfl_v}) and obtained
\begin{equation}\label{L_fl_ttt}
\begin{aligned}
\!\!\mathbb{\ell }_{\mathrm{afl}}\! &=\!\! \sum_{i=1}^{\mathcal{N} } \left [\! (1\!+\!\alpha)(1\!-\!\mathbf{P}^{i}_{\mathrm{t}})^{\gamma_{\mathrm{d}} +1}\!+\!\frac{1}{2} (1\!-\!\mathbf{P}^{i}_{\mathrm{t}})^{\gamma_{\mathrm{d}} +2}\!+\!... \right ],
\end{aligned}
\end{equation}

Next, we derive Eq.~(\ref{L_bce_ttt}) of $\mathbb{\ell }_{\mathrm{bce}}$, which yields:
\begin{equation}\label{L_dbce}
\begin{aligned}
- \frac{\partial \mathbb{\ell }_{\mathrm{bce}}}{\partial \mathbf{P}_{\mathrm{t}}} \!&=\! -\!\!\! \sum_{i=1}^{\mathcal{N} } \nu _{i}\! =\!\! \sum_{i=1}^{\mathcal{N} } \left [1\!\! +\!\! \left(1\!\! - \!\!\mathbf{P}^{i}_{\mathrm{t}} \right)^{1}\!\! + \!\!\left (1\!\!-\!\!\mathbf{P}^{i}_{\mathrm{t}} \right)^{2} \!\! + \!...  \right ],	
\end{aligned}
\end{equation}
where $\nu _{i}=\partial \mathbb{\ell }_{\mathrm{bce}} / \partial \mathbf{P}^{i}_{\mathrm{t}}$ is the gradient of BCE for each pixel in the sample. The first term of the $\nu _{i}$ in Eq.~(\ref{L_dbce}) is always $1$, regardless of the difficulty of the pixel, which indicates that the gradient of $\mathbb{\ell }_{\mathrm{bce}}$ is ``difficulty-equal'', treating all pixels equally.
Because of this property, $\mathbb{\ell }_{\mathrm{bce}}$ can perform competitively in pixels with uniform difficulty distribution, but often performs poorly when the difficulty pixels are imbalanced (see Fig.~\ref{fig:fig3}, the visualization of BCE~\cite{yi2004automated}).

\begin{figure*}[tp]
\centering
\includegraphics[width=0.9\linewidth]{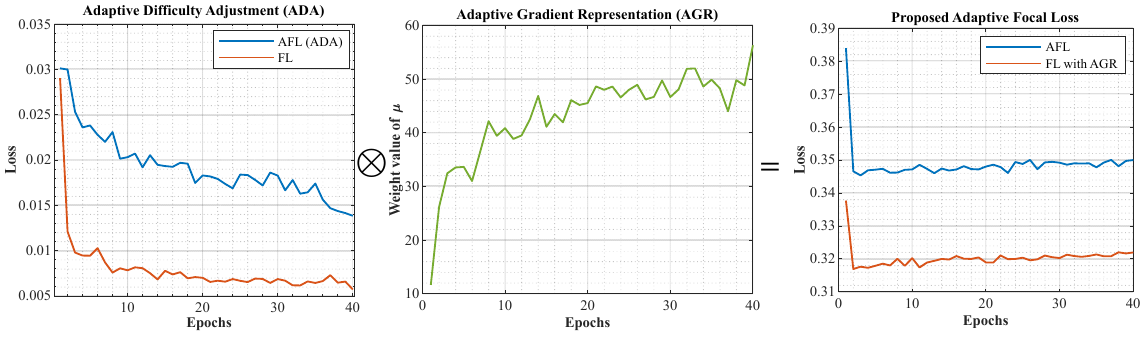}	
\caption{A visualization plot of the gradient correction of FL~\cite{lin2017focal} by the ADA and AGR is shown} in this paper. The left figure is the FL after the ADA adjustment. The middle figure is the gradient adjustment value generated by AGR. The right figure is the FL and AFL being adjusted.
\label{fig:fig6}
\end{figure*}

Interestingly, when deriving Eq.~(\ref{L_fl_ttt}) we can obtain,
\begin{equation}\label{L_fl_gr}
\begin{aligned}
- \frac{\partial \mathbb{\ell }_{\mathrm{afl}}}{\partial \mathbf{P}_{\mathrm{t}}}\!\! &=\!\!\! \sum_{i=1}^{\mathcal{N}} (1\!\!-\! \mathbf{P}^{i}_{\mathrm{t}})^{\gamma_{\mathrm{d}}} \!\! \left [(1\!\!+\!\!\alpha)(1\!\!+\!\!\gamma_{\mathrm{d}}) \! +\! (1\!+\!\!\frac{\gamma_{\mathrm{d}}}{2})(1\!\!-\!\!\mathbf{P}^{i}_{\mathrm{t}})\! + \! ... \right ]\!\!,
\end{aligned}
\end{equation}
In contrast to $\mathbb{\ell }_{\mathrm{bce}}$, the gradients of $\mathbb{\ell }_{\mathrm{afl}}$ and $\mathbb{\ell }_{\mathrm{fl}}$ are ``difficulty-oriented'', so the gradient swamping of $\mathbb{\ell }_{\mathrm{afl}}$ refer to the ambiguous pixels being swamped by the hard pixels.

In view of the above analysis, we expect that $\mathbb{\ell }_{\mathrm{afl}}$ can have the ability of $\mathbb{\ell }_{\mathrm{fl}}$ to classify hard and easy pixels while taking into account the focus on ambiguous pixels like $\mathbb{\ell }_{\mathrm{bce}}$.
Observing Eq.~(\ref{L_dbce}) and Eq.~(\ref{L_fl_gr}), we can see that the gradient of $\mathbb{\ell }_{\mathrm{afl}}$ can be approximated as the  $\mathbb{\ell }_{\mathrm{bce}}$ gradient discarding the first $\gamma_{\mathrm{d}}$ terms. Therefore, there are two ways to make the $\mathbb{\ell }_{\mathrm{afl}}$'s gradient approximate to $\mathbb{\ell }_{\mathrm{bce}}$: (1) adding the discarding gradients of $\mathbb{\ell }_{\mathrm{bce}}$ to make $\mathbb{\ell }_{\mathrm{afl}}$ to approximate $\mathbb{\ell }_{\mathrm{bce}}$; (2) multiplying by an Adaptive Gradient Representation (AGR) factor $\mu$ that forces the $\mathbb{\ell }_{\mathrm{afl}}$ approximate to $\mathbb{\ell }_{\mathrm{bce}}$. 
Unfortunately, the scheme (1) relies on adding the missing terms $\sum_{i=1}^{\mathcal{N} } \sum_{j=1}^{\gamma_{\mathrm{d}}} (1-\mathbf{P}^{i}_{\mathrm{t}})^{j+1}$ in Eq.~(\ref{L_fl_gr}), which still retains the stiff ``difficulty-equal'' properties of $\mathbb{\ell }_{\mathrm{bce}}$. Therefore, our primary focus is on exploring the gradient approximation scheme of (2).

Based on this conjecture, we summarized the gradient of $\mathbb{\ell }_{\mathrm{bce}}$ with respect to that of $\mathbb{\ell }_{\mathrm{afl}}$ at the gradient level. 
Then, we find that the adjustment of $\mathbb{\ell }_{\mathrm{afl}}$ to the $\mathbb{\ell }_{\mathrm{bce}}$ is mainly focused on the vertical direction of the gradient, \textit{e.g.}, in Eq.~(\ref{L_bjz1}),
\begin{equation}\label{L_bjz1}
\small
\begin{aligned}
-\frac{\partial \mathbb{\ell }_{\mathrm{afl}}}{\partial \mathbf{P}_{\mathrm{t}}} = \sum_{i=1}^{\mathcal{N} }  (1-\mathbf{P}^{i}_{\mathrm{t}})^{\gamma_{\mathrm{d}}} \begin{bmatrix} 1\phantom{(1-\mathbf{P}^{i}_{\mathrm{t}})} &\!\!\!\!\! \!\!\!\!\! + \gamma_{\mathrm{d}}(1 + \alpha + \frac{\alpha}{\gamma_{\mathrm{d}}})  & \!\! \!\! \!\!\!\! \!\! \!\!\\
+(1-\mathbf{P}^{i}_{\mathrm{t}})\phantom{^{2}}  &\!\!\!\!\! \! + \frac{\gamma_{\mathrm{d}}}{2}(1-\mathbf{P}^{i}_{\mathrm{t}})\phantom{\gamma_{\mathrm{d}}\gamma_{\mathrm{d}}2}  &\!\! \!\! \!\!\!\! \!\! \!\!\\
+(1-\mathbf{P}^{i}_{\mathrm{t}})^{2} &\!\!\!\!\! \! + \frac{\gamma_{\mathrm{d}}}{3}(1-\mathbf{P}^{i}_{\mathrm{t}})^{2}\phantom{\gamma_{\mathrm{d}}\gamma2} \!\!& \!\! \!\! \!\!\!\! \!\! \!\!\\
+...\phantom{(-\mathbf{P}^{i}_{\mathrm{t}})^{2}} &\!\!\!\!\! \! + ...\phantom{(-\mathbf{P}^{i}_{\mathrm{t}})^{2}-\mathbf{P}^{i}_{2}2222} \!\! \!\!\!\! \!\! \!\!
\end{bmatrix},
\end{aligned}
\end{equation}
In Eq.~(\ref{L_bjz1}), the left column of the polynomial is $\nu _{i}$. The right column of the polynomial is the correction gradient of vertical generated by $(1-\mathbf{P}^{i}_{\mathrm{t}})^{\gamma_{\mathrm{d}}}$, which we define as $\nabla _{\mathcal{B}}$.

Since there is only a quantitative difference between $\nabla _{\mathcal{B}}$ and $\nu _{i}$, we consider $\forall \delta, -\nu _{i} = \delta \nabla _{\mathcal{B}}$, where $\delta \in [0, 1]$. Based on the above analysis, we can obtain Eq.~(\ref{L_gn}),
\begin{equation}\label{L_gn}
\begin{aligned}
\frac{\partial \mathbb{\ell }_{\mathrm{afl}}}{\partial \mathbf{P}_{\mathrm{t}}} &= \sum_{i=1}^{\mathcal{N} }  (1-\mathbf{P}^{i}_{\mathrm{t}})^{\gamma_{\mathrm{d}}} \left [\nu _{i} + \delta \nabla _{\mathcal{B}} \right ]\\
&= \sum_{i=1}^{\mathcal{N} }  (1-\mathbf{P}^{i}_{\mathrm{t}})^{\gamma_{\mathrm{d}}} \left [(1 + \delta \gamma_{\mathrm{d}})\nu _{i} \right ],
\end{aligned}
\end{equation}

The model classification reaches optimality under the condition that $\sum_{i=1}^{\mathcal{N} } (1-\mathbf{P}^{i}_{\mathrm{t}})^{\gamma_{\mathrm{d}}} \nu _{i}$ satisfies each $(1-\mathbf{P}^{i}_{\mathrm{t}})^{\gamma_{\mathrm{d}}} \in \sum_{i=1}^{\mathcal{N} } (1-\mathbf{P}^{i}_{\mathrm{t}})^{\gamma_{\mathrm{d}}} = 0$, and each $\nu _{i} \in \sum_{i=1}^{\mathcal{N} } \nu _{i}=1$.
Then, based on Chebyshev's inequality~\cite{heinig1991chebyshev}, we can obtain
\begin{equation}\label{L_che}
\begin{aligned}
\sum_{i=1}^{\mathcal{N} } (1-\mathbf{P}^{i}_{\mathrm{t}})^{\gamma_{\mathrm{d}}} \nu _{i} = \tfrac{1}{\mathcal{N} } \sum_{i=1}^{\mathcal{N} } (1-\mathbf{P}^{i}_{\mathrm{t}})^{\gamma_{\mathrm{d}}}\sum_{i=1}^{\mathcal{N} } \nu _{i},
\end{aligned}
\end{equation}

Collating Eqs.~(\ref{L_dbce}),~(\ref{L_gn}), and~(\ref{L_che}), we can get,
\begin{equation}\label{L_gn2}
\begin{aligned}
\frac{\partial  \mathbb{\ell }_{\mathrm{bce}}} {\partial \mathbf{P}_{\mathrm{t}}} =  \frac{\partial \mathbb{\ell }_{\mathrm{afl}}/\partial \mathbf{P}_{\mathrm{t}}}{\frac{1}{\mathcal{N} } \sum_{i=1}^{\mathcal{N} }  (1-\mathbf{P}^{i}_{\mathrm{t}})^{\gamma_{\mathrm{d}}} (1+ \delta  \gamma_{\mathrm{d}})},
\end{aligned}
\end{equation}

Observing Eq.~(\ref{L_gn}) and Eq.~(\ref{L_gn2}), the gradient of $\mathbb{\ell }_{\mathrm{afl}}$ can be represented as an approximate total gradient of $\mathbb{\ell }_{\mathrm{bce}}$ by introducing $\mu$, where
\begin{equation}\label{beta}
\small
\begin{aligned}
\mu  =\frac{\mathcal{N}}{\sum_{i=1}^{\mathcal{N} } {(1-\mathbf{P}^{i}_{\mathrm{t}})^{\gamma_{\mathrm{d}}} (1+\delta  \gamma_{\mathrm{d}}})}  .
\end{aligned}
\end{equation}

Based on the above analysis, $\mu$ allows the gradient of the $\mathbb{\ell }_{\mathrm{afl}}$ to be balanced between ``difficulty-equal'' and ``difficulty-oriented'', which can increase the weight of ambiguity pixels in the training process, thus making the $\mathbb{\ell }_{\mathrm{afl}}$ notice the ambiguity pixels rather than focusing more on extremely hard pixels.

\subsubsection{Adaptive Focal Loss}

In summary, $\mathbb{\ell }_{\mathrm{afl}}$ is as follows, 
\begin{equation}\label{L_BNFL_nfl}
\begin{aligned}
\mathbb{\ell }_{\mathrm{afl}} &=\!\! \sum_{i=1}^{\mathcal{N} }  \left [ - \mu (1-\mathbf{P}^{i}_{\mathrm{t}})^{\gamma_{\mathrm{d}}} \mathrm{log}(\mathbf{P}^{i}_{\mathrm{t}}) + \alpha (1\!-\!\mathbf{P}^{i}_{\mathrm{t}})^{\gamma_{\mathrm{d}}+1} \right ].
\end{aligned}
\end{equation}

As shown in Fig.~\ref{fig:fig6}, ADA can not only make the $\mathbb{\ell }_{\mathrm{afl}}$ focus more on ambiguity pixels but also produce a larger margin compared to FL~\cite{lin2017focal}, which indicates that AFL can adjust the training strategy based on different sample distributions and learning situations. On the other hand, the AGR allows the $\mathbb{\ell }_{\mathrm{afl}}$ to be balanced between ``difficulty-equal'' and ``difficulty-oriented'', which can increase the weight of ambiguity pixels, yielding solving the gradient swamping. In addition, the ADA can not only make the model focus on ambiguous pixels but is also better at maintaining the stability of the model when using AGR directly on FL. This is reflected by the fact that the loss value of FL drifts more when AGR is used, while AFL does not in the right sub-figure of Fig.~\ref{fig:fig6}.

Moreover, it can be easily obtained that when the values of $\mu $ and $\gamma_{\mathrm{d}}$ are $0$, the proposed AFL is equivalent to PL~\cite{leng2022polyloss}; when $\mu $, $\gamma_{\mathrm{a}}$, and $\alpha$ are $0$, AFL is equivalent to FL~\cite{lin2017focal}; when $\mu$, $\gamma_{\mathrm{d}}$, and $\alpha$ are $0$, AFL is equivalent to BCE~\cite{yi2004automated}. This flexible quality allows AFL to achieve higher accuracy when compared to other approaches.

\subsection{Model Optimization} \label{sec:MP}

\textbf{Mask Loss.} The proposed AFL and Dice Loss~\cite{milletari2016v} are used as the mask loss in this work.
Here,
\begin{equation}\label{L_mask}
\begin{aligned}
\mathbb{\ell }_{\mathrm{mask}} (\hat{\mathbf{y}}^{i}_{\mathrm{p}}, \mathbf{y}^{i}_{\mathrm{p}}) =  \lambda _{\mathrm{afl}} \mathbb{\ell }_{\mathrm{afl}} (\hat{\mathbf{y}}^{i}_{\mathrm{p}}, \mathbf{y}^{i}_{\mathrm{p}}) + \lambda _{\mathrm{dice}} \mathbb{\ell }_{\mathrm{dice}} (\hat{\mathbf{y}}^{i}_{\mathrm{p}}, \mathbf{y}^{i}_{\mathrm{p}}),
\end{aligned}
\end{equation}
where $\hat{\mathbf{y}}^{i}_{\mathrm{p}}$ is the click-predicted instances permuted according to the optimal permutation $\hat{\mathbb{\sigma}} \in \mathbf{S}_{n_{q}}$.

\textbf{Click Loss.} The cross-entropy loss is used as the click loss, and the click loss $\mathbb{\ell }_{\mathrm{cli}} (\hat{c}_{m}, c_{m})$ is intended to compute classification confidence for each click, which in turn ensures that each click can be computed efficiently.

\textbf{Total Loss.}
Ultimately, the total loss can be expressed as,
\begin{equation}\label{L_total}
\begin{aligned}
\mathbb{L }_{\mathrm{total}} \left(\hat{\mathbf{y}}^{i}_{\mathrm{p}}, \mathbf{y}^{i}_{\mathrm{p}} \right ) =\!\! \sum_{i=1}^{\mathcal{N}} \lambda _{\mathrm{mask}}  \mathbb{\ell }_{\mathrm{mask}} \left (\hat{\mathbf{y}}^{i}_{\mathrm{p}}, \mathbf{y}^{i}_{\mathrm{p}} \right ) \!+\! \lambda _{\mathrm{cli}}\mathbb{\ell }_{\mathrm{cli}} (\hat{c}_{m}, c_{m}).
\end{aligned}
\end{equation}
For predictions that match the $\hat{\mathbf{y}}^{i}_{\mathrm{p}}$ and $0.1$ for ``unclick'', \textit{i.e.}, predictions that do not match any GT.

\section{Experiments} \label{sec:experiments}

\subsection{Datasets} \label{sec:datasets}

We evaluate the proposed AdaptiveClick using the following well-recognized datasets widely used in IIS tasks:

(1) SBD~\cite{majumder2019SBD}: Comprising $8,498$ images for training and $2,857$ images for testing, this dataset is characterized by its scene diversity  and is frequently utilized for IIS tasks;

(2) COCO-LIVS~\cite{sofiiuk2022reviving}: This dataset comprises $118$K training images (with $1.2$M instances), which is widely adopted due to its diverse class distributions~\cite{sofiiuk2022reviving,liu2022simpleclick,lin2014microsoft,zhou2023interactive}.

(3) GrabCut~\cite{rother2004grabcut}: The GrabCut dataset contains $50$ images, which have relatively simple appearances and have also been commonly used for evaluating the performance of different IIS methods~\cite{sofiiuk2022reviving,liu2022simpleclick,lin2014microsoft,zhou2023interactive};

(4) Berkeley~\cite{mcguinness2010berkeley}: This dataset includes $100$ images, sharing some small object images with GrabCut~\cite{rother2004grabcut}. It poses challenges for IIS models due to these similarities;

(5) DAVIS~\cite{perazzi2016DIVAS}: Initially designed for video image segmentation, the $50$ videos are divided into $345$ frames for testing. 
The dataset features images with high-quality masks.

(6) Pascal VOC~\cite{everingham2010VOC}: 
It is composed of $1449$ images ($3427$ instances). We assess segmentation performance on this validation set, as in~\cite{chen2021CDNet,lin2020first,liu2022simpleclick};

(7) ssTEM~\cite{gerhard2013segmented}: The ssTEM dataset contains two image stacks, each with $20$ medical images. We use the same stack as in~\cite{liu2022simpleclick, liu2022pseudoclick} to evaluate model effectiveness;

(8) BraTS~\cite{baid2021rsna}: The BraTS dataset includes $369$ Magnetic Resonance Images (MRI) volumes. We test our model on the same $369$ slices as in~\cite{liu2022simpleclick, liu2022pseudoclick};

(9) OAIZIB~\cite{ambellan2019automated}: OAIZIB dataset consists of $507$ MRI volumes. We test our model on the same $150$ slices ($300$ instances) as in~\cite{liu2022isegformer,liu2022simpleclick}.

\begin{table*} [htbp]

\centering
\renewcommand\arraystretch{1.0}
\setlength\tabcolsep{4pt}
\caption{Comparison in NoC85 and NoC90 between AdaptiveClick and state-of-the-art methods trained on Augmented VOC~\cite{everingham2010VOC} and SBD~\cite{majumder2019SBD} datasets and tested on GrabCut~\cite{rother2004grabcut}, Berkeley~\cite{mcguinness2010berkeley}, SBD~\cite{majumder2019SBD}, DAVIS~\cite{perazzi2016DIVAS}, and Pascal VOC~\cite{everingham2010VOC} datasets.
Top results within a group are indicated in \underline{underline} and the overall top results in \textbf{bold}.}
\resizebox{\linewidth}{!}{\begin{tabular}{l|c|c|cc|cc|cc|cc|cc|cc}
\toprule [2pt]
\multirow{2}{*}{Method}       & \multirow{2}{*}{Backbone} & \multirow{2}{*}{Train Data} & \multicolumn{2}{c}{GrabCut} & \multicolumn{2}{c}{Berkeley} & \multicolumn{2}{c}{SBD} & \multicolumn{2}{c}{DAVIS}  & \multicolumn{2}{c}{Pascal VOC} & \multicolumn{2}{c}{Average}\\
\cline{4-15}
&                           &                             & NoC85 & NoC90      & NoC85      & NoC90  & NoC85 & NoC90   & NoC85 & NoC90  & NoC85 & NoC90   & NoC85 & NoC90\\ 		
\toprule [1pt]
DIOS~\cite{xu2016DIOS}            $_{\mathrm{CVPR2016}}$   & FCN                         & Augmented VOC               & -   & $6.04$           & -   & $8.65$        & -   & -          & -  & $12.58$   & $6.88$  & -   & - & -        \\
FCANet~\cite{lin2020first}        $_{\mathrm{CVPR2020}}$     & ResNet101                 & Augmented VOC               & -      & $2.08$                    & -  & $3.92$    & -              & -   & -   & $7.57$    & $2.69$      & -   & - & -  \\
Latent Diversity~\cite{li2018LD}        $_{\mathrm{CVPR 2018}}$ & VGG-19          & SBD                         & $3.20$    & $4.79$              & -     & -   & $7.41$   & $10.78$          & $5.05$   & $9.57$   & - & -    & - & -       \\
BRS~\cite{jang2019BRS}                  $_{\mathrm{CVPR2019}}$    & DenseNet                         & SBD                         & $2.60$    & $3.60$        & -        & $5.08$     & $6.59$   & $9.78$           & $5.58$   & $8.24$   & - &-   & - & -        \\
IA-SA~\cite{kontogianni2020IA-SA}         $_{\mathrm{ECCV2020}}$         & ResNet101                  & SBD                         & -    & $3.07$         & -      & $4.94$     & -   & -           & $5.16$   & -    & - & -   & - & -       \\
f-BRS-B~\cite{sofiiuk2020f-brs}         $_{\mathrm{CVPR2020}}$         & ResNet50                  & SBD                         & $2.50$    & $2.98$     & -          & $4.34$     & $5.06$   & $8.08$           & $5.39$   & $7.81$       &- & -    & - & -   \\
CDNet~\cite{chen2021CDNet}              $_{\mathrm{CVPR2021}}$       & ResNet34                  & SBD                         & $1.86$   & $2.18$           & $1.95$    & $3.27$     & $5.18$   & $7.89$           & $5.00$   & $6.89$     & $3.61$ & $4.51$   & 3.52 &  {4.95}      \\
RITM~\cite{sofiiuk2022reviving}         $_{\mathrm{ICIP2022}}$            & HRNet18                   & SBD                         & $1.76$   & $2.04$      & 1.87         & $3.22$     & $3.39$   & $5.43$           & $4.94$   & $6.71$    & $2.51$  & $3.03$    &  {2.89} &  {4.09}         \\
PseudoClick~\cite{liu2022pseudoclick}   $_{\mathrm{ECCV2022}}$         & HRNet18                   & SBD                         & $1.68$      & $2.04$    & $1.85$     & $3.23$     & $3.38$      & $5.40$            & $4.81$ & $6.57$   & $\underline{2.34}$  & $\underline{2.74}$    &  {2.81} &  {4.00}  \\	
FocalClick~\cite{chen2022focalclick}    $_{\mathrm{CVPR2022}}$           & SegF-B0     & SBD       & $1.66$   & $1.90$     &-           & $3.14$     & $4.34$   & $6.51$           & $5.02$   & $7.06$       &-&-   &  {-} &  {-}     \\
FcousCut~\cite{lin2022focuscut}    $_{\mathrm{CVPR2022}}$             & ResNet101                 & SBD                         & $1.46$   & $1.64$         & $1.81$      & $3.01$     & $3.40$    & $5.31$           & $4.85$   & $6.22$      & - &-   &  {-} &  {-}      \\
GPCIS~\cite{zhou2023interactive}    $_{\mathrm{CVPR2023}}$             & SegF-B0                 & SBD                         & $1.60$   & $1.76$         & $1.84$      & $2.70$     & $4.16$    & $6.28$           & $4.45$   & $6.04$       & - & -   &  {-} &  {-}    \\
FCFI~\cite{wei2023focused}  $_{\mathrm{CVPR2023}}$  & ResNet101 & SBD & $1.64$ & $1.80$ &- & $2.84$ & $\underline{3.26}$ & $5.35$ & $4.75$ & $6.48$ & - & - &  {-} &  {-} \\
 {EMC~\cite{du2023efficient}}  $_{\mathrm{ {CVPR2023}}}$  &  {HRNet18} &  {SBD} & $ {1.74}$ & $ {1.84}$ &  {-}  & $ {3.03}$ & $ {3.38}$ & $ {5.51}$ & $ {5.05}$ & $ {6.71}$ & $ {2.37}$ & - &  {-} &  {-} \\
 {FDRN~\cite{zeng2023feature}} $_{\mathrm{ {ACM MM2023}}}$  &  {SegF-B0} &  {SBD} &  {1.58} &  {1.78} & - &  {3.08} &  {4.18} &  {6.20} &  {4.78} &  {6.66} &  {-} &  {-} &  {-} &  {-} \\
SimpleClick~\cite{liu2022simpleclick}    $_{\mathrm{ {ICCV2023}}}$             & ViT-B                 & SBD                         & $\underline{1.40}$   & $\underline{1.54}$         & $\underline{1.44}$      & $\underline{2.46}$     & $3.28$    & $\underline{5.24}$           & $\underline{4.10}$   & $\underline{5.48}$       & $2.38$ & $2.81$  & \underline{2.52} & \underline{3.51}     \\
\toprule [1pt]
\rowcolor[gray]{.9} 
\textbf{AdaptiveClick}   (ours)         & ViT-B                   & SBD    & $\textbf{1.38}$  & $\textbf{1.46}$   & $\textbf{1.38}$  & $\textbf{2.18}$   & $\textbf{3.22}$  & $\textbf{5.22}$ & $\textbf{4.00}$   & $\textbf{5.14}$   & $\textbf{2.25}$    & $\textbf{2.66}$   & \textbf{2.45} & \textbf{3.33} \\	
\toprule [2pt]         
\end{tabular}}


\label{tab:tab5}
\end{table*}

\begin{table*} [htbp]
\centering
\renewcommand\arraystretch{1.0}
\setlength\tabcolsep{5pt}
\caption{Comparison in NoC85 and NoC90 between AdaptiveClick and state-of-the-art methods trained on the COCO-LVIS dataset~\cite{sofiiuk2022reviving} and tested on GrabCut~\cite{rother2004grabcut}, Berkeley~\cite{mcguinness2010berkeley}, SBD~\cite{majumder2019SBD}, DAVIS~\cite{perazzi2016DIVAS}, and Pascal VOC~\cite{everingham2010VOC} datasets. $^{\dagger }$ denotes the result is from SEEM.}
\resizebox{\linewidth}{!}{\begin{tabular}{l|c|c|cc|cc|cc|cc|cc|cc}			
\toprule [2pt]
\multirow{2}{*}{Method} & \multirow{2}{*}{Backbone} & \multirow{2}{*}{Train Data} & \multicolumn{2}{c}{GrabCut} & \multicolumn{2}{c}{Berkeley} & \multicolumn{2}{c}{SBD} & \multicolumn{2}{c}{DAVIS}  & \multicolumn{2}{c}{Pascal VOC} & \multicolumn{2}{c}{ {Average}}\\		
\cline{4-15}
&            &           & NoC85 & NoC90 & NoC85 & NoC90   & NoC85 & NoC90   & NoC85 & NoC90   & NoC85 & NoC90   &  {NoC85} &  {NoC90}   \\ 		
\toprule [1pt]	
f-BRS-B~\cite{sofiiuk2020f-brs}     $_{\mathrm{CVPR2020}}$       & HRNet32            & COCO-LVIS       & $1.54$   & $1.69$   & -            & $2.44$     & $4.37$   & $7.26$           & $5.17$   & $6.50$  & - &-    &  {-} &  {-}          \\
CDNet~\cite{chen2021CDNet}     $_{\mathrm{ICCV 2021}}$      & ResNet34             & COCO-LVIS        & $1.40$   & $1.52$  & $1.47$             & $2.06$      & $4.30$   & $7.04$   & $4.27$   & $5.56$      & $2.74$  & $3.30$  &  {2.84} &  {3.90}  \\
RITM~\cite{sofiiuk2022reviving}     $_{\mathrm{ICIP2022}}$          & HRNet32    & COCO-LVIS  & $1.46$   & $1.56$  & $1.43$       & $2.10$      & $3.59$   & $5.71$           & $4.11$   & $5.34$      & $2.19$  & $2.57$  &  {2.56} &  {3.46}     \\
FocalClick~\cite{chen2022focalclick}  $_{\mathrm{CVPR2022}}$      & SegF-B0                   & COCO-LVIS                   & $1.40$   & $1.66$ & $1.59$         & $2.27$      & $4.56$     & $6.86$   & $4.04$           & $5.49$   & $2.97$ & $3.52$  &  {2.91} &  {3.96}            \\
FocalClick~\cite{chen2022focalclick}  $_{\mathrm{CVPR2022}}$      & SegF-B3               & COCO-LVIS                   & $1.44$   & $1.50$        & $1.55$        & $1.92$     & $3.53$   & $5.59$           & $3.61$   & $4.90$    & $2.46$  & $2.88$   &  {2.52} &  {3.56}       \\
PseudoClick~\cite{liu2022pseudoclick}  $_{\mathrm{ECCV2022}}$      & HRNet32                   & COCO-LVIS                   & $1.36$      & $1.50$        & $1.40$        & $2.08$     & $3.38$      & $5.54$           & $3.79$   & $5.11$   & $1.94$  & $2.25$    &  {2.37} &  {3.30}        \\
FCFI~\cite{wei2023focused}  $_{\mathrm{CVPR2023}}$  & HRNet18 & COCO-LVIS & $1.38$ & $1.46$ & - & $1.96$ & $3.63$ & $5.83$ & $3.97$ & $5.16$ & - & - &  {-} &  {-} \\
 {EMC~\cite{du2023efficient}}  $_{\mathrm{ {CVPR2023}}}$  &  {SegF-B3} &  {COCO-LVIS} & $ {1.42}$ & $ {1.48}$ &  {-} & $ {2.35}$ & $ {3.44}$ & $ {5.57}$ & $ {4.49}$ & $ {5.69}$ & $ {2.23}$ & - &  {-} &  {-} \\
 {FDRN~\cite{zeng2023feature}} $_{\mathrm{ {ACM MM2023}}}$  &  {SegF-B3} &  {COCO-LVIS} &  {1.42} &  {1.44} & - &  {1.80} &  {3.74} &  {5.57} &  {3.55} &  {4.90} &  {-} &  {-} &  {-} &  {-} \\
DynaMITe~\cite{RanaMahadevan23DynaMITe} $_{\mathrm{ {ICCV2023}}}$     & SegF-B3                   & COCO-LVIS                  & $1.48$    & $1.58$        & $1.34$     & $1.97$           & $3.81$   & $6.38$          & $3.81$   & $5.00$     & -    & -  &  {-} &  {-}  \\	
$^{\dagger }$ SAM~\cite{kirillov2023segany}  $_{\mathrm{ {ICCV2023}}}$     & ViT-B                   & COCO-LVIS                  & -    & -        & -     & -           & $6.50$   & $9.76$          & -   & -     & $3.30$    & $4.20$  &  {-} &  {-}  \\
$^{\dagger }$ SEEM~\cite{zou2023segment}  $_{\mathrm{ {NeurIPS2023}}}$     & DaViT-B                   & COCO-LVIS                  & -    & -        & -     & -           & $6.67$   & $9.99$          & -   & -     & $3.41$    & $4.33$  &  {-} &  {-}  \\
InterFormer  ~\cite{huang2023interformer}  $_{\mathrm{ {ICCV2023}}}$     & ViT-B                   & COCO-LVIS                  & $1.38$    & $1.50$        & $1.99$     & $3.14$           & $3.78$   & $6.34$          & $4.10$   & $6.19$     & -    & - &  {-} &  {-}  \\	
InterFormer  ~\cite{huang2023interformer}  $_{\mathrm{ {ICCV2023}}}$     & ViT-L                   & COCO-LVIS                  & $\textbf{1.26}$    & $\textbf{1.36}$        & $1.61$     & $2.53$           & $3.25$   & $5.51$          & $4.54$   & $5.21$     & -    & -  &  {-} &  {-}  \\
 {iCMFormer~\cite{li2023interactive}  $_{\mathrm{ {ICCVW2023}}}$}  &  {ViT-B}    &  {COCO-LVIS}   & $ {1.42}$    & $ {1.52}$        & $ {1.40}$     & $ {1.86}$           & $ {3.29}$   & $ {5.30}$          & $3.40$   & $ {5.06}$     &  {-} &  {-}   &  {-} &  {-}  \\	
SimpleClick~\cite{liu2022simpleclick}    $_{\mathrm{ {ICCV2023}}}$             & ViT-H                 & COCO-LVIS                         & $1.38$   & $1.50$          & $\underline{1.36}$     & $1.75$     & $\underline{2.85}$    & $\underline{4.70}$   & $3.41$   & $\underline{4.78}$  & $\underline{1.76}$    & $\underline{1.98}$ & \underline{2.15} & \underline{2.94} \\
 {VTMR~\cite{fang2023varianceinsensitive}  $_{\mathrm{ {AAAI2024}}}$}  &  {SegF-B3}    &  {COCO-LVIS}   & $ {1.38}$    & $ {1.42}$        & $ {1.44}$     & $\underline{1.72}$           & $ {3.55}$   & $ {5.53}$          & $\underline{3.26}$   & $ {4.82}$     &  {-} &  {-}   &  {-} &  {-}  \\	
\toprule [1pt]	
\rowcolor[gray]{.9} 
\textbf{AdaptiveClick } (ours)          & ViT-H                   & COCO-LVIS                   & $\underline{1.32}$    & $\underline{1.38}$               & $\textbf{1.32}$      & $\textbf{1.64}$   & $\textbf{2.84}$     & $\textbf{4.68}$      & $\textbf{3.19}$   & $\textbf{4.60}$     & $\textbf{1.74}$    & $\textbf{1.96}$  & \textbf{2.08} & \textbf{2.85}  \\	
\toprule [2pt]
\end{tabular}}

\label{tab:tab6}
\end{table*}

\subsection{Implementation Details} \label{Implementation details}

For data embedding, given the clicks with previous mask $\left \{\mathbf{C}^1, \mathbf{C}^2, ..., \mathbf{C}^m, \textbf{M}_{\mathrm{pre}} \right \}$ and $\mathbf{I}$ to obtain encode features $\mathbf{C}_\mathrm{t}$ and $\mathbf{I}_{\mathrm{t}}$ via patch embedding, then they are fused  like~\cite{sofiiuk2022reviving,liu2022simpleclick,chen2022focalclick}. For feature encoding, we use ViT-Base (ViT-B) and ViT-Huge (ViT-H) as our backbone. 
For feature decoding, the designed PMMD is used to encode the fusion feature and then sent to the CAMD, where CAMD consists of three transformer blocks, \textit{i.e.}, $L=3$ ($9$ layers in total), and $\mathbf{Q}$ is set to $10.0$. For post-processing, we multiply the click and mask confidence to obtain the final confidence matrix, then generate the $\mathbf{P}_{\mathrm{n}}$ as done in~\cite{cheng2022masked, cheng2021per}. In addition, $\mathbb{L }_{\mathrm{total}}$ is added to each CAMD layer to guarantee the performance of the model.

For AdaptiveClick training, we train for $60$ epochs on the SBD~\cite{majumder2019SBD} and the COCO-LVIS dataset~\cite{sofiiuk2022reviving}; the initial learning rate is set to $5 \times 10^{-5}$, and then reduced by $10$ times in the $40$-th epoch. The image is cropped to size $448 \times 448$ pix.
$\lambda_{\mathrm{cli}}$ is $2$, $\lambda_{\mathrm{mask}}$ is $1$, $\lambda_{\mathrm{afl}}$ is $5$, and $\lambda_{\mathrm{dice}}$ is $5$ in this work. Adam is used to optimize the training with the parameters $\beta_1=0.9$ and $\beta_2=0.999$. The batch size is $48$ for training ViT-B and $24$ for training ViT-H, respectively. All experiments are trained and tested on NVIDIA Titan RTX $6000$ GPUs.

In addition, to assess the robustness and generalization of AFL, we have incorporated it into state-of-the-art IIS methods and ensured consistency in other parameter settings. The hyperparameters $\gamma$, $\delta$, and $\alpha$ for the proposed AFL are fixed at $2$, $0.4$, and $1.0$, respectively, throughout the training process.

\subsection{Evaluation Metrics} \label{sec:exp_metrics}

Following previous works~\cite{sofiiuk2022reviving, chen2022focalclick, lin2020first}, we adopt the combination of Number of Clicks (NoC) and Intersection over Union (IoU) at $0.85$ (NoC85) and $0.9$ (NoC90) as evaluation metrics. Here, NoC90 indicates that the IoU of the mask obtained by the IIS model is $90$ when NoC is $k$. In particular, the IIS task sets the maximum NoC to $20$ (samples with NoC $>20$ are considered failures). Therefore, the lower the value of NoC85 and NoC90, the better. We also use the average IoU given $k$ clicks (mIoU@$k$) as an evaluation metric to measure the segmentation quality given a fixed number of clicks.

\subsection{Comparison Methods}  \label{sec:exp_compared_method}

We validate our approach against state-of-the-art IIS methods and loss functions, specifically:

\textbf{IIS Methods.} IIS methods including DIOS~\cite{xu2016DIOS}, Latent diversity~\cite{li2018LD}, BRS~\cite{jang2019BRS}, f-BRS-B~\cite{sofiiuk2020f-brs}, IA-SA~\cite{kontogianni2020IA-SA}, FCA-Net~\cite{lin2020first}, PseudoClick~\cite{liu2022pseudoclick}, CDNet~\cite{chen2021CDNet}, RITM~\cite{sofiiuk2022reviving}, FocalClick~\cite{chen2022focalclick}, FocusCut~\cite{lin2022focuscut}, GPCIS~\cite{zhou2023interactive}, SimpleClick~\cite{liu2022simpleclick}, FCFI~\cite{wei2023focused}, InterFormer~\cite{huang2023interformer}, DynaMITe~\cite{RanaMahadevan23DynaMITe}, SAM~\cite{kirillov2023segany}, SEEM~\cite{zou2023segment}, EMC~\cite{du2023efficient}, FDRN~\cite{zeng2023feature}, iCMFormer~\cite{li2023interactive}, and VTMR~\cite{fang2023varianceinsensitive}, all of which achieve competitive results.

\begin{figure*}[tbp]
\centering
\includegraphics[width=0.8\textwidth]{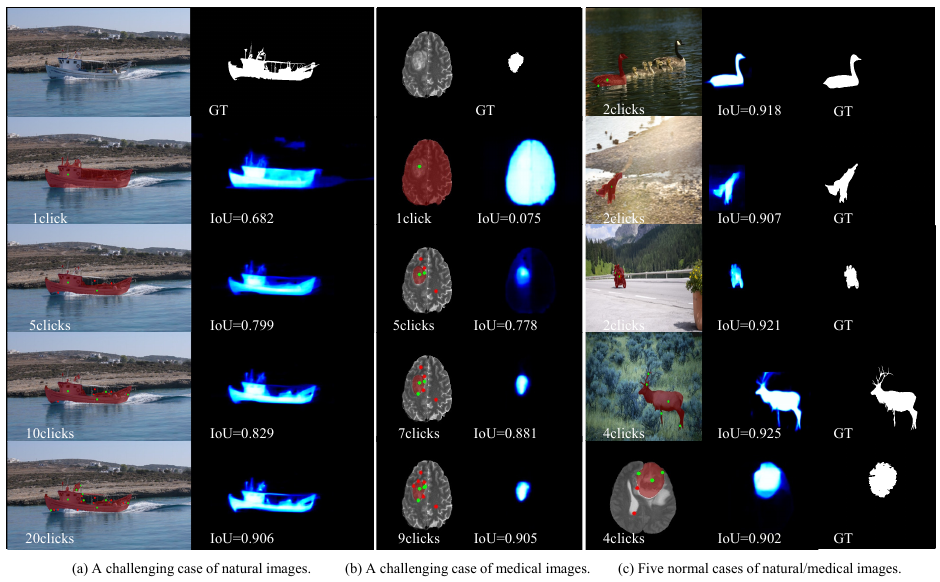}
\caption{Segmentation results on natural and medical datasets.
The backbone is ViT-B trained on the SBD dataset~\cite{majumder2019SBD}. The probability maps are shown in blue; the masks are overlaid in red on the original images. The clicks are shown as green (positive click) or red (negative click) dots on the image.
\label{fig:fig7}}
\end{figure*}

\textbf{Loss Functions.} We compare the AFL with other loss functions, including BCE~\cite{yi2004automated}, Soft IoU~\cite{rahman2016optimizing}, FL~\cite{lin2017focal}, NFL~\cite{sofiiuk2019nfocal}, WBCE~\cite{pihur2007weighted}, Balanced CE~\cite{xie2015holistically}, and PL~\cite{leng2022polyloss} losses, which are widely used in IIS and semantic segmentation tasks~\cite{jadon2020survey}. Specifically, in FL and NFL, the value of $\gamma$ is $2$, and the balanced weight is $0.5$. For PL, the value of $\gamma$ is $2$, the balanced weight is $0.5$, and the value of $\alpha$ is $1$. Moreover, the positive weight for WBCE is $3$, while the positive and negative weights for the Balanced CE are $3$ and $1$, respectively.

\subsection{Model Analysis} \label{sec:model_analysis}

\begin{table} [tbp!]
\centering
\renewcommand\arraystretch{1.14}
\setlength\tabcolsep{1pt}
\caption{Comparison in NoC85 and NoC90 between AdaptiveClick and state-of-the-art methods trained on SBD~\cite{majumder2019SBD} and tested on ssTEM~\cite{rother2004grabcut}, BraTS~\cite{mcguinness2010berkeley}, and OAIZIB~\cite{mcguinness2010berkeley}.}
\resizebox{\linewidth}{!}{\begin{tabular}{l|c|c|cc|cc|cc}			
\toprule [2pt]
\multirow{2}{*}{Method} & \multirow{2}{*}{Backbone} & \multirow{2}{*}{Train Data} & \multicolumn{2}{c}{ssTEM} & \multicolumn{2}{c}{BraTS} & \multicolumn{2}{c}{OAIZIB} \\		
\cline{4-9}
&       &    & NoC85 & NoC90 & NoC85 & NoC90   & NoC85 & NoC90      \\ \toprule [1pt]
CDNet~\cite{chen2021CDNet}     $_{\mathrm{ICCV 2021}}$          & ResNet34           & SBD    & $11.10$   & $14.65$  & $17.07$     & $18.86$      & $19.56$   & $19.95$     \\
RITM~\cite{sofiiuk2022reviving}     $_{\mathrm{ICIP2022}}$          & HRNet18                   & SBD                  & $\underline{3.71}$   & $5.68$  & $\textbf{8.47}$             & $\textbf{12.59}$      & $\underline{17.70}$   & $\underline{19.95}$                   \\
SimpleClick~\cite{liu2022simpleclick}    $_{\mathrm{ {ICCV2023}}}$             & ViT-B          & SBD          & $3.78$	& $\underline{5.21}$	& $\underline{9.93}$	& $\underline{13.90}$	& $18.44$	& $19.90$     \\
\toprule [1pt]
\rowcolor[gray]{.9} 
\textbf{AdaptiveClick} (ours)  & ViT-B        & SBD        & $\textbf{3.17}$    & $\textbf{4.56}$               & $10.79$      & $14.39$   & $\textbf{17.43}$           & $\textbf{19.62}$    \\
\toprule [2pt]
\end{tabular}}
\label{tab:tab7}
\end{table}

\subsubsection{Comparison with State-of-the-Art Methods} \label{sec:exp_comparison_soa}

TABLE~\ref{tab:tab5} and TABLE~\ref{tab:tab6} empirically analyze the performance of AdaptiveClick and state-of-the-art methods using SBD and COCO-LIVS as training datasets.
Likewise, TABLE~\ref{tab:tab7} and TABLE~\ref{tab:tab8} display the performance of AdaptiveClick and relevant published methods in medical images, respectively.

\textbf{Evaluation on the Natural Images.} 
In TABLE~\ref{tab:tab5} and~\ref{tab:tab6}, we conduct a comparative analysis between AdaptiveClick and state-of-the-art IIS methods. Firstly, we provide an experimental comparison of AdaptiveClick using ViT-B and ViT-H as the backbones, along with the proposed AFL as the loss function. AdaptiveClick achieves an average accuracy improvement of $0.074$, $0.174$, and $0.07$, $0.09$ on SBD and COCO-LIVS, respectively, when compared to existing state-of-the-art methods on NoC85 and NoC90. This underscores the robustness and superior accuracy of AdaptiveClick in addressing interaction ambiguity and gradient swamping challenges. 

Simultaneously, we have conducted a statistical analysis of the average NoC85 and NoC90 for various IIS methods. The results demonstrate that AdaptiveClick outperforms all comparison methods, achieving more accurate segmentation with fewer clicks.

\begin{table} [tbp!]
\centering
\renewcommand\arraystretch{1.0}
\setlength\tabcolsep{1pt}
\caption{Comparison in NoC85 and NoC90 between AdaptiveClick and state-of-the-art methods trained on COCO-LVIS~\cite{sofiiuk2022reviving} and tested on ssTEM~\cite{rother2004grabcut}, BraTS~\cite{mcguinness2010berkeley}, and OAIZIB~\cite{mcguinness2010berkeley}.}
\resizebox{\linewidth}{!}{\begin{tabular}{l|c|c|cc|cc|cc}			
\toprule [2pt]
\multirow{2}{*}{Method} & \multirow{2}{*}{Backbone} & \multirow{2}{*}{Train Data} & \multicolumn{2}{c}{ssTEM} & \multicolumn{2}{c}{BraTS} & \multicolumn{2}{c}{OAIZIB} \\		
\cline{4-9}
&      &   & NoC85 & NoC90 & NoC85 & NoC90   & NoC85 & NoC90  \\ 		
\toprule [1pt]
CDNet~\cite{chen2021CDNet}     $_{\mathrm{ICCV 2021}}$          & ResNet34           & COCO-LVIS    & $4.15$   & $8.45$  & $10.51$     & $14.80$      & $17.42$   & $19.81$     \\
RITM~\cite{sofiiuk2022reviving}     $_{\mathrm{ICIP2022}}$          & HRNet32   & COCO-LVIS     & $\textbf{2.74}$   & $\textbf{4.06}$  & $7.56$     & $11.24$      & $15.89$   & $19.27$  \\
FocalClick~\cite{chen2022focalclick}  $_{\mathrm{CVPR2022}}$      & SegF-B3               & COCO-LVIS            & $3.95$   & $5.05$  & $7.17$             & $11.19$      & $\underline{12.93}$   & $19.23$          \\
SimpleClick~\cite{liu2022simpleclick}    $_{\mathrm{ {ICCV2023}}}$             & ViT-H         & COCO-LVIS       & $4.27$	& $5.45$	& $\underline{6.73}$	& $\underline{10.27}$	& $14.93$	&$\underline{18.95}$ \\
\toprule [1pt]
\rowcolor[gray]{.9} 
\textbf{AdaptiveClick}  (ours)   & ViT-H   & COCO-LVIS  & $\underline{3.95}$    & $\underline{4.97}$      & $\textbf{6.51}$      & $\textbf{9.77}$   & $\textbf{12.65}$           & $\textbf{16.69}$     \\
\toprule [2pt]
\end{tabular}}
\label{tab:tab8}
\end{table}

\textbf{Evaluation on Medical Images.}  
The results of AdaptiveClick and state-of-the-art IIS methods for medical images are presented in TABLEs~\ref{tab:tab7} and~\ref{tab:tab8}. 
It is noteworthy that prevalent IIS methods generally demonstrate suboptimal performance when applied to medical image segmentation, attributable to two key factors. 
Firstly, the majority of IIS methods have not undergone training on specialized medical datasets, despite the introduction of such datasets in some studies for training purposes. 
Secondly, these IIS methods often display limited generalization capabilities. 
In contrast, AdaptiveClick consistently achieves high accuracy across various medical datasets, highlighting the method's versatility. In addition, Fig.~\ref{fig:fig7} is the result of some challenging and normal examples, and our AdaptiveClick can segment the specified object in the image across both natural and medical datasets.

\begin{table} [tbp!]
\centering
\renewcommand\arraystretch{1.0}
\setlength\tabcolsep{4pt}
\caption{Computation comparison for model Parameters, FLOPs, GPU memory, and Speed with different state-of-the-art IIS methods.}
\resizebox{\linewidth}{!}{\begin{tabular}{l|cccc}			
\toprule [2pt]
 {Method (backbone, size)} &  {Params/M} &  {FLOPs/G} &  {Mem/G} & $ {\downarrow }$  {SPC/ms} \\
\cline{1-5}
\toprule [1pt]
 {RITM (HRNet32, 400)$_{\mathrm{ICIP 2022}}$}~\cite{sofiiuk2022reviving}  & $ {30.95}$    & $ {83.12}$    & $ {0.50}$      & $ {54}$\\
 {iSegFormer (Swin-L, 400)$_{\mathrm{MICCAI 2022}}$}~\cite{liu2022isegformer}  & $ {195.90}$    & $ {302.78}$    & $ {2.14}$      & $ {44}$\\
 {FocalClick (SegF-B3, 256)$_{\mathrm{CVPR 2022}}$}~\cite{chen2022focalclick}  & $ {45.66}$    & $ {24.75}$    & $ {0.32}$      & $ {53}$\\
 {FocusCut (ResNet101, 384)$_{\mathrm{CVPR 2022}}$}~\cite{lin2022focuscut}  & $ {59.35}$    & $ {100.76}$    & $ {0.89}$      & $ {355}$\\
 {SimpleClick (ViT-B, 448)$_{\mathrm{ICCV 2023}}$}~\cite{liu2022simpleclick}  & $ {96.46}$    & $ {169.78}$    & $ {0.87}$      & $ {54}$\\
 {InterFormer (ViT-B, 512)$_{\mathrm{ICCV 2023}}$}~\cite{huang2023interformer}  & $ {120.39}$    & $ {533.70}$    & $ {1.40}$      & $ {360}$\\
 {iCMFormer (ViT-B, 512)$_{\mathrm{ICCVW 2023}}$}~\cite{li2023interactive} & $ {124.81}$ & $ {297.54}$ &  {-}  &  {78} \\
\toprule [1pt]  
\rowcolor[gray]{.9} 
 {AdaptiveClick (ViT-B, 448) (ours)}  & $ {116.41}$    & $ {269.81}$    & $ {1.28}$      & $ {74}$\\
\toprule [2pt]     
\end{tabular}}

\label{tab:tab15}
\end{table}

\textbf{Computational Analysis.} TABLE~\ref{tab:tab15} provides computation statistics results for state-of-the-art IIS methods. As in~\cite{liu2022simpleclick}, we evaluate the AdaptiveClick and compare it with existing methods on GrabCut~\cite{rother2004grabcut}. The experimental results reveal that, although the proposed mask-adaptive method introduces some computational overhead, with a memory consumption of $1.28$G and a running speed of $74$ms, it still meets the requirements for real-time annotation. This demonstrates that AdaptiveClick not only effectively addresses the interaction ambiguity but also enhances segmentation accuracy while meeting real-time demands.

\subsubsection{Effectiveness of Adaptive Focal Loss}
\label{sec:exp_fl}

The performance of different loss functions on the SBD~\cite{majumder2019SBD} and COCO-LIVS~\cite{sofiiuk2022reviving} datasets is illustrated in TABLE~\ref{tab:tab1} and TABLE~\ref{tab:tab2}, respectively. 
TABLEs~\ref{tab:tab3} and~\ref{tab:tab4} report the experimental results of embedding AFL into existing methods on SBD and COCO-LIVS training datasets, respectively.

\begin{figure*}[htbp]
\centering
\includegraphics[width=0.76\linewidth]{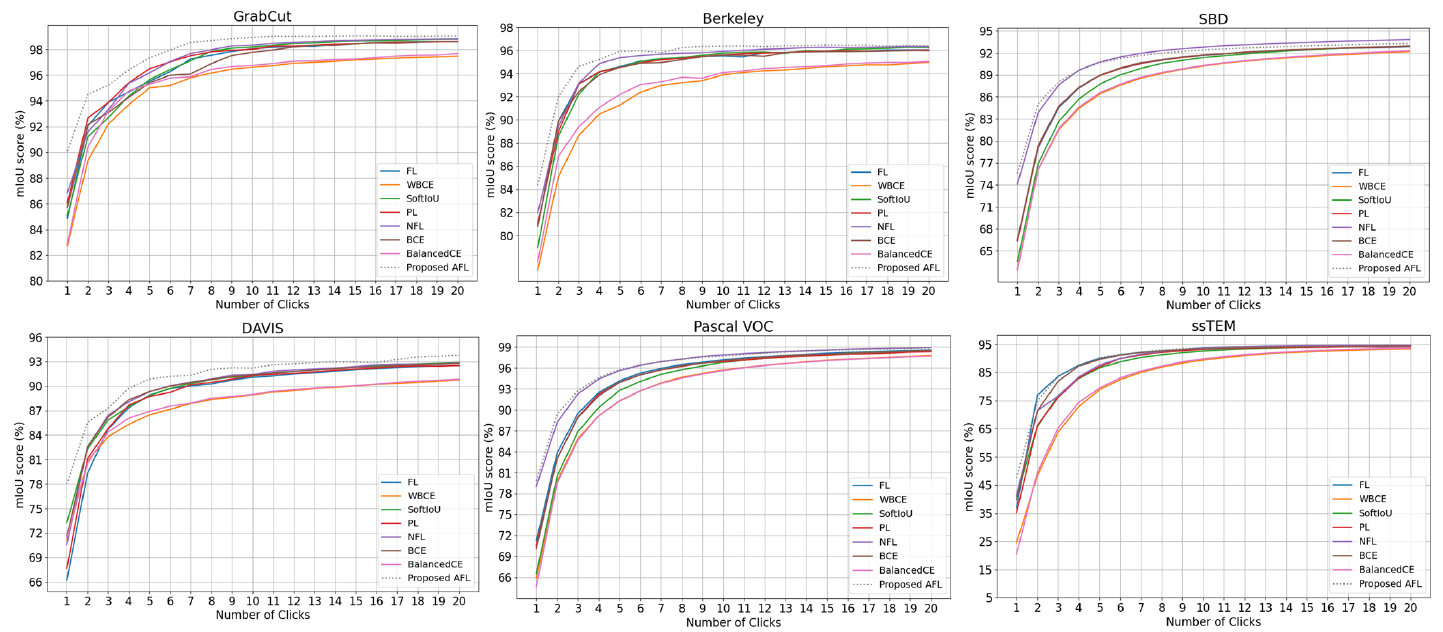}	
\caption{The convergence analysis between AFL and state-of-the-art loss functions. AdaptiveClick is used as the baseline and SBD~\cite{majumder2019SBD} as the training dataset. The test datasets are GrabCut~\cite{rother2004grabcut}, Berkeley~\cite{mcguinness2010berkeley}, SBD~\cite{majumder2019SBD}, DAVIS~\cite{perazzi2016DIVAS}, Pascal VOC~\cite{everingham2010VOC}, and ssTEM~\cite{rother2004grabcut}, respectively.
The metric is the mean IoU given $k$ clicks. Overall, our models require fewer clicks for a given accuracy level.
The methods from top to bottom are FL~\cite{lin2017focal}, WBCE~\cite{pihur2007weighted}, Soft IoU~\cite{rahman2016optimizing},  PL~\cite{leng2022polyloss}, NFL~\cite{sofiiuk2019nfocal}, BCE~\cite{yi2004automated}, Balanced CE~\cite{xie2015holistically}, and AFL, respectively. Our AdaptiveClick in general requires fewer clicks for a given accuracy level.}
\label{fig:fig8}
\end{figure*}

\begin{figure*}[tbp]
\centering
\includegraphics[width=0.76\linewidth]{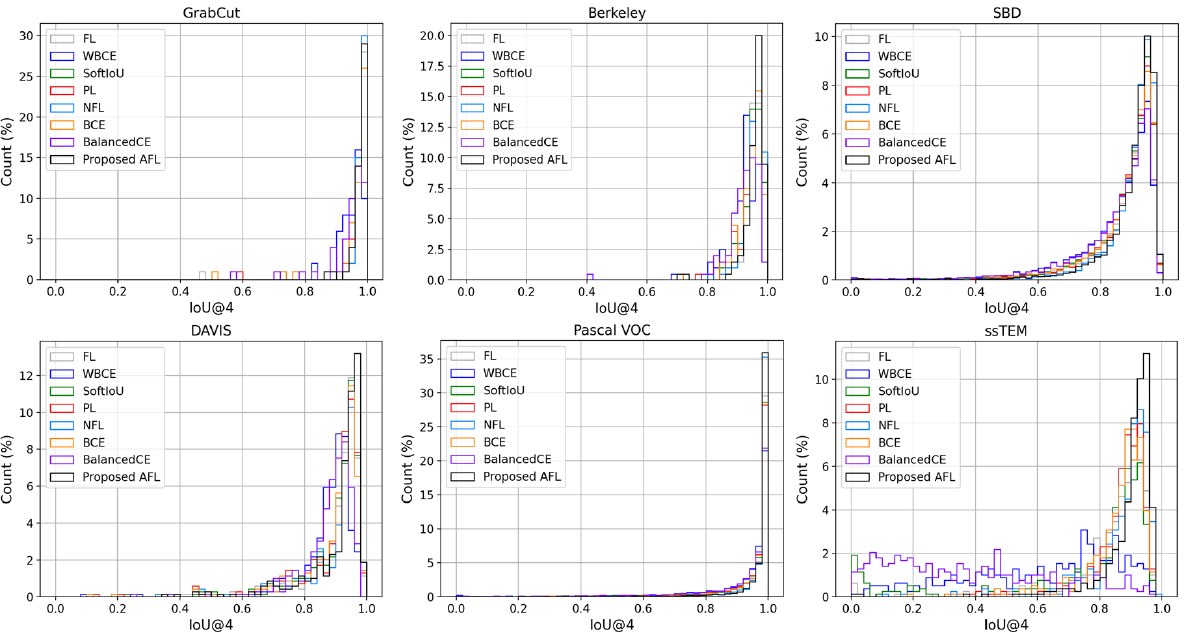}	
\caption{The histogram analysis between AFL and state-of-the-art loss functions. In this experiment, AdaptiveClick is used as the baseline and COCO-LVIS~\cite{sofiiuk2022reviving} as the training dataset. The test datasets are GrabCut~\cite{rother2004grabcut}, Berkeley~\cite{mcguinness2010berkeley}, SBD~\cite{majumder2019SBD}, DAVIS~\cite{perazzi2016DIVAS}, Pascal VOC~\cite{everingham2010VOC}, and ssTEM~\cite{rother2004grabcut}, respectively.The methods from top to bottom are FL~\cite{lin2017focal}, WBCE~\cite{pihur2007weighted}, Soft IoU~\cite{rahman2016optimizing}, PL~\cite{leng2022polyloss}, NFL~\cite{sofiiuk2019nfocal}, BCE~\cite{yi2004automated}, Balanced CE~\cite{xie2015holistically}, and the proposed AFL, respectively. Given $k=4$ clicks, AdaptiveClick obtains the best results (more instances with a higher IoU) than comparison loss functions.}
\label{fig:fig9}
\end{figure*}

\textbf{Comparison with State-of-the-Art Loss Functions.} In TABLE~\ref{tab:tab1}, the efficacy of AFL is demonstrated through an average improvement of $0.014{\sim}0.07$ and $1.136{\sim}2.504$ on NoC85 and NoC90, respectively (SBD training dataset). This improvement is observed in comparison to BCE, WBCE, Balanced CE, FL, NFL, and PL. 
The results suggest that, as analyzed in Eq.~(\ref{L_dbce}) and Eq.~(\ref{L_fl_gr}), both BCE and FL-based losses exhibit a certain degree of gradient swamping. The proposed AFL proves effective in mitigating this issue.
When comparing the experimental results of AFL with WBCE, Balanced CE, and Soft IoU, it becomes evident that while these losses address the pixel imbalance problem in segmentation, they do not excel in the IIS task, due to inadequate resolution of the more profound gradient swamping issue. This observation is supported by Fig.~\ref{fig:fig8}, which illustrates the mean IoU performance for each loss function. The figure affirms that AFL consistently enhances performance across all six datasets.

\begin{table} [tbp]
\centering
\setlength\tabcolsep{1.0pt}
\renewcommand\arraystretch{1.0}
\caption{Evaluation of AdaptiveClick of ViT-B trained on the SBD~\cite{majumder2019SBD} with different loss functions. We report NoC85 and NoC90 on GrabCut~\cite{rother2004grabcut}, Berkeley~\cite{mcguinness2010berkeley}, SBD~\cite{majumder2019SBD}, DAVIS~\cite{perazzi2016DIVAS}, and Pascal VOC~\cite{everingham2010VOC} datasets.}
\resizebox{\linewidth}{!}{\begin{tabular}{l|cc|cc|cc|cc|cc}
\toprule [2pt]
\multirow{2}{*}{Method}   & \multicolumn{2}{c}{GrabCut} & \multicolumn{2}{c}{Berkeley} & \multicolumn{2}{c}{SBD} & \multicolumn{2}{c}{DAVIS}  & \multicolumn{2}{c}{Pascal VOC} \\	
\cline{2-11}			
 & NoC85 & NoC90  &  NoC85 & NoC90 & NoC85 & NoC90  & NoC85 & NoC90  & NoC85 & NoC90\\
\toprule [1pt]
BCE~\cite{yi2004automated}  $_{\mathrm{ISIMP 2004}}$  & $1.64$    & $1.78$               & $1.69$        & $2.99$   & $4.06$           & $6.39$     & $4.55$ & $6.08$  & $2.85$ & $3.37$           \\
WBCE~\cite{pihur2007weighted}   $_{\mathrm{Bioinf. 2007}}$  & $1.66$   & $2.58$               & $2.25$     & $5.20$   & $5.18$           & $8.00$    & $5.43$  & $9.28$  & $3.39$ & $4.12$            \\
Balanced CE~\cite{xie2015holistically}  $_{\mathrm{ICCV 2015}}$  & $1.62$   & $1.84$               & $1.88$     & $3.59$   & $4.55$           & $7.00$    & $5.57$ & $7.35$ & $3.44$  & $4.18$             \\
Soft IoU~\cite{rahman2016optimizing}    $_{\mathrm{ISVC 2016}}$   & $1.62$   & $1.99$               & $1.78$     & $2.29$   & $4.55$           & $6.48$    & $4.69$ & $6.32$ &$3.13$ & $3.71$            \\
FL~\cite{lin2017focal}      $_{\mathrm{ICCV 2017}}$   & $1.62$   & $1.84$               & $1.69$     & $2.82$   & $4.11$           & $6.51$    & $4.94$  & $6.50$ & $2.81$ & $3.35$            \\
NFL~\cite{sofiiuk2019nfocal}  $_{\mathrm{ICCV 2019}}$   & $\underline{1.40}$   & $\underline{1.50}$               & $\textbf{1.32}$     & $\underline{2.24}$   & $\underline{3.25}$           & $\underline{5.28}$    & $\underline{4.03}$   & $\underline{5.24}$   & $\underline{2.30}$  & $\underline{2.75}$      \\
PL~\cite{xie2015holistically}  $_{\mathrm{ICLR 2022}}$  & $1.46$   & $1.70$               & $1.68$     & $2.87$   & $4.09$           & $6.42$    & $4.70$ & $6.43$  & $2.85$  & $3.39$           \\
\toprule [1pt]
\rowcolor[gray]{.9} 
\textbf{AFL} (ours) & $\textbf{1.38}$    & $\textbf{1.46}$               & $\underline{1.38}$      & $\textbf{2.18}$   & $\textbf{3.22}$           & $\textbf{5.22}$   & $\textbf{4.00}$     & $\textbf{5.14}$    & $\textbf{2.25}$  & $\textbf{2.66}$     \\
\toprule [2pt]  
\end{tabular}}
\label{tab:tab1}
\end{table}

\begin{table} [tbp]
\centering
\setlength\tabcolsep{1.0pt}
\renewcommand\arraystretch{1.0}
\caption{Evaluation of AdaptiveClick of ViT-B trained on the COCO-LVIS~\cite{sofiiuk2022reviving} with different loss functions. We report NoC85 and NoC90 on GrabCut~\cite{rother2004grabcut}, Berkeley~\cite{mcguinness2010berkeley}, SBD~\cite{majumder2019SBD}, DAVIS~\cite{perazzi2016DIVAS}, and Pascal VOC~\cite{everingham2010VOC} datasets.}
\resizebox{\linewidth}{!}{\begin{tabular}{l|cc|cc|cc|cc|cc}
\toprule [2pt]
\multirow{2}{*}{Method}  & \multicolumn{2}{c}{GrabCut} & \multicolumn{2}{c}{Berkeley} & \multicolumn{2}{c}{SBD} & \multicolumn{2}{c}{DAVIS}  & \multicolumn{2}{c}{Pascal VOC} \\	
\cline{2-11}			
&  NoC85 & NoC90  & NoC85  & NoC90 & NoC85 & NoC90  & NoC85 & NoC90  & NoC85 & NoC90\\
\toprule [1pt]
BCE~\cite{yi2004automated}  $_{\mathrm{ISIMP 2004}}$  & $1.54$ & $	1.72$	& $1.79$ & $	2.89$  & $	4.30$ &$	6.74$	&$4.29$	& $5.76$	&$2.71$	& $3.19$	        \\
WBCE~\cite{pihur2007weighted}   $_{\mathrm{Bioinf. 2007}}$  & $ 1.60$	&$1.88$	&$2.15$	&$4.29$	&$5.60$	&$8.61$	&$5.26$	&$8.79$	&$3.32$	&$4.03$			        \\
Balanced CE~\cite{xie2015holistically}  $_{\mathrm{ICCV 2015}}$  & $1.74$	&$2.10$	&$2.13$	&$4.32$	&$5.53$	&$8.58$	&$4.90$	&$8.02$	&$3.24$	&$3.93$ \\
Soft IoU~\cite{rahman2016optimizing}    $_{\mathrm{ISVC 2016}}$  & $1.62$   & $	1.70$               & $	1.64$     & $	2.63$   & $	4.19$      & $	6.66$    & $4.19$       & $	5.60$   & $	2.69$  & $	3.18$   \\
FL~\cite{lin2017focal}      $_{\mathrm{ICCV 2017}}$  & $1.54$	&$1.60$	&$1.58$	&$2.62$	&$4.19$	&$6.60$	&$4.12$	&$5.60$	&$2.64$	&$3.12$ \\
NFL~\cite{sofiiuk2019nfocal}  $_{\mathrm{ICCV 2019}}$  & $\textbf{1.34}$    & $\underline{1.54}$               & $\underline{1.55}$      & $\underline{1.92}$   & $\underline{3.34}$           & $\underline{5.48}$   & $\underline{3.66}$     & $\underline{4.93}$    & $\underline{2.06}$  & $\underline{2.39}$ \\
PL~\cite{xie2015holistically}  $_{\mathrm{ICLR 2022}}$  & $\underline{1.50}$   & $1.66$               & $1.66$     & $2.85$   & $4.26$           & $6.74$    & $4.32$    & $5.68$  & $2.71$  & $3.19$        \\
\toprule [1pt]
\rowcolor[gray]{.9} 
\textbf{AFL} (ours)  & $\textbf{1.34}$    & $\textbf{1.48}$               & $\textbf{1.40}$      & $\textbf{1.83}$   & $\textbf{3.29}$           & $\textbf{5.40}$   & $\textbf{3.39}$     & $\textbf{4.82}$    & $\textbf{2.03}$ & $\textbf{2.31}$  \\
\toprule [2pt]  
\end{tabular}}
\label{tab:tab2}
\end{table}

\begin{table*} [htbp]
\centering
\renewcommand\arraystretch{1.0}
\setlength\tabcolsep{8pt}
\caption{Results in NoC85 and NoC90 for state-of-the-art methods with and without the proposed AFL. In these experiments, we trained on the SBD dataset~\cite{majumder2019SBD} and tested on GrabCut~\cite{rother2004grabcut}, Berkeley~\cite{mcguinness2010berkeley}, SBD~\cite{majumder2019SBD}, DAVIS~\cite{perazzi2016DIVAS}, and Pascal VOC~\cite{everingham2010VOC} datasets. It should be noted that AdaptiveClick + NFL indicates the results obtained by using AdaptiveClick but using NFL~\cite{sofiiuk2019nfocal} as the training loss. AdaptiveClick, on the other hand, indicates the results obtained by using both the framework of this paper and the AFL as loss functions. Top results within a group are indicated in \textcolor{red}{red}.}
\resizebox{\linewidth}{!}{\begin{tabular}{l|c|c|cc|cc|cc|cc|cc}
\toprule [2pt]
\multirow{2}{*}{Method}       & \multirow{2}{*}{Backbone} & \multirow{2}{*}{Train Data} & \multicolumn{2}{c}{GrabCut} & \multicolumn{2}{c}{Berkeley} & \multicolumn{2}{c}{SBD} & \multicolumn{2}{c}{DAVIS}  & \multicolumn{2}{c}{Pascal VOC}\\
\cline{4-13}
&                           &                             & NoC85 & NoC90      & NoC85      & NoC90  & NoC85 & NoC90   & NoC85 & NoC90  & NoC85 & NoC90  \\ 		
\toprule [1pt]
CDNet~\cite{chen2021CDNet}              $_{\mathrm{CVPR2021}}$       & ResNet34                  & SBD                         & $1.86$   & $2.18$           & $1.95$    & $3.27$     & $5.18$   & $7.89$           & $5.00$   & $6.38$     & - & -         \\
\rowcolor[gray]{.9} 
\textbf{CDNet} (ours)       & ResNet34                   & SBD                  & $\textcolor{red}{1.82}$    & $\textcolor{red}{2.14}$               & $\textcolor{red}{1.86}$      & $\textcolor{red}{2.88}$   & $\textcolor{red}{4.53}$           & $\textcolor{red}{7.15}$   & $\textcolor{red}{4.78}$  & $\textcolor{red}{6.27}$   & - & -   \\	
\toprule [1pt]
RITM~\cite{sofiiuk2022reviving}         $_{\mathrm{ICIP2022}}$            & HRNet18                   & SBD                         & $1.76$   & $2.04$      & 1.87         & $3.22$     & $\textcolor{red}{3.39}$   & $5.43$           & $4.94$   & $6.71$    & $2.51$  & $3.03$        \\
\rowcolor[gray]{.9} 
\textbf{RITM} (ours)            & HRNet18                   & SBD                         & $\textcolor{red}{1.60}$   & $\textcolor{red}{1.86}$         & $\textcolor{red}{1.72}$     & $\textcolor{red}{2.93}$   & $3.40$           & $\textcolor{red}{5.42}$   & $\textcolor{red}{4.72}$ & $\textcolor{red}{6.15}$     & $\textcolor{red}{2.41}$  & $\textcolor{red}{2.92}$   \\
\toprule [1pt]
FocalClick~\cite{chen2022focalclick}    $_{\mathrm{CVPR2022}}$           & SegF-B0               & SBD                         & $1.66$   & $1.90$     &-           & $3.14$     & $\textcolor{red}{4.34}$   & $6.51$           & $5.02$   & $7.06$       &-&-       \\
\rowcolor[gray]{.9} 
\textbf{FocalClick} (ours)            & SegF-B0                   & SBD                         & $\textcolor{red}{1.64}$   & $\textcolor{red}{1.84}$           &-    & $\textcolor{red}{3.12}$     & $4.36$   & $\textcolor{red}{6.44}$           & $\textcolor{red}{4.79}$   & $\textcolor{red}{6.50}$     & -  & -       \\
\toprule [1pt]
FcousCut~\cite{lin2022focuscut}    $_{\mathrm{CVPR2022}}$              & ResNet50                  & SBD                         & $1.60$    & $1.78$      & $1.85$         & $3.44$     & $3.62$   & $5.66$           & $5.00$      & $6.38$          & - & -    \\
\rowcolor[gray]{.9} 
\textbf{FcousCut} (ours)            & ResNet50                 & SBD                         & $\textcolor{red}{1.54}$   & $\textcolor{red}{1.72}$               & $\textcolor{red}{1.79}$     & $\textcolor{red}{3.34}$    & $\textcolor{red}{3.50}$           & $\textcolor{red}{5.62}$   & $\textcolor{red}{4.82}$   & $\textcolor{red}{6.28}$   &- 	&-        \\
\toprule [1pt]
SimpleClick~\cite{liu2022simpleclick}    $_{\mathrm{ {ICCV2023}}}$             & ViT-B                 & SBD                         & $1.40$   & $1.54$         & $\textcolor{red}{1.44}$      & $2.46$     & $3.28$    & $5.24$           & $\textcolor{red}{4.10}$   & $\textcolor{red}{5.48}$       & $2.38$ & $2.81$      \\
\rowcolor[gray]{.9} 
\textbf{SimpleClick} (ours)              & ViT-B                  & SBD                  & $\textcolor{red}{1.36}$   & $\textcolor{red}{1.52}$ & $\textcolor{red}{1.44}$  & $\textcolor{red}{2.34}$        & $\textcolor{red}{3.10}$       & $\textcolor{red}{5.04}$   & $\textcolor{red}{4.10}$     & $5.51$  & $\textcolor{red}{2.23}$  & $\textcolor{red}{2.63}$ \\		
\toprule [1pt]
AdaptiveClick + NFL          & ViT-B                   & SBD                   & $1.40$    & $1.50$               & $\textcolor{red}{1.32}$      & $2.24$   & $3.25$           & $5.28$   & $4.03$     & $5.24$    & $2.30$  & $2.75$   \\	
\rowcolor[gray]{.9} 
\textbf{AdaptiveClick} (ours)          & ViT-B                   & SBD                   & $\textcolor{red}{1.38}$    & $\textcolor{red}{1.46}$               & $1.38$      & $\textcolor{red}{2.18}$   & $\textcolor{red}{3.22}$           & $\textcolor{red}{5.22}$   & $\textcolor{red}{4.00}$     & $\textcolor{red}{5.14}$    & $\textcolor{red}{2.25}$  & $\textcolor{red}{2.66}$    \\		
\toprule [2pt]         
\end{tabular}}

\label{tab:tab3}
\end{table*}

\begin{table*} [htbp]
\centering
\renewcommand\arraystretch{1.0}
\setlength\tabcolsep{7pt}
\caption{Results in NoC85 and NoC90 for state-of-the-art methods with and without the proposed AFL. In these experiments, we trained on COCO-LVIS~\cite{sofiiuk2022reviving} and tested on the GrabCut~\cite{rother2004grabcut}, Berkeley~\cite{mcguinness2010berkeley}, SBD~\cite{majumder2019SBD}, DAVIS~\cite{perazzi2016DIVAS}, and Pascal VOC~\cite{everingham2010VOC} datasets.}
\resizebox{\linewidth}{!}{\begin{tabular}{l|c|c|cc|cc|cc|cc|cc}			
\toprule [2pt]
\multirow{2}{*}{Method} & \multirow{2}{*}{Backbone} & \multirow{2}{*}{Train Data} & \multicolumn{2}{c}{GrabCut} & \multicolumn{2}{c}{Berkeley} & \multicolumn{2}{c}{SBD} & \multicolumn{2}{c}{DAVIS}  & \multicolumn{2}{c}{Pascal VOC}\\		
\cline{4-13}
&                           &                             & NoC85 & NoC90 & NoC85 & NoC90   & NoC85 & NoC90   & NoC85 & NoC90   & NoC85 & NoC90     \\ 
\toprule [1pt]
RITM~\cite{sofiiuk2022reviving}     $_{\mathrm{ICIP2022}}$          & HRNet32                   & COCO-LVIS                   & $1.46$   & $1.56$  & $1.43$             & $2.10$      & $3.59$   & $5.71$           & $4.11$   & $5.34$      & $2.19$  & $2.57$      \\
\rowcolor[gray]{.9} 
\textbf{RITM} (ours)       & HRNet32                   & COCO-LVIS                   & $\textcolor{red}{1.42}$   & $\textcolor{red}{1.51}$          & $\textcolor{red}{1.40}$     & $\textcolor{red}{1.98}$      & $\textcolor{red}{3.48}$   & $\textcolor{red}{5.64}$           & $\textcolor{red}{3.69}$   & $\textcolor{red}{5.26}$      & $\textcolor{red}{2.10}$  & $\textcolor{red}{2.54}$\\
\toprule [1pt]
FocalClick~\cite{chen2022focalclick}  $_{\mathrm{CVPR2022}}$      & SegF-B3               & COCO-LVIS                   & $1.44$   & $1.50$        & $1.55$        & $1.92$     & $3.53$   & $5.59$           & $3.61$   & $4.90$    & $2.46$  & $2.88$         \\
\rowcolor[gray]{.9} 
\textbf{FocalClick} (ours)        & SegF-B3                   & COCO-LVIS                   & $\textcolor{red}{1.40}$    & $\textcolor{red}{1.44}$               & $\textcolor{red}{1.52}$      & $\textcolor{red}{1.87}$   & $\textcolor{red}{3.51}$          & $\textcolor{red}{5.49}$  & $\textcolor{red}{3.57}$     & $\textcolor{red}{4.85}$    & $\textcolor{red}{2.44}$ 	& $\textcolor{red}{2.86}$     \\
\toprule [1pt]
SimpleClick~\cite{liu2022simpleclick}    $_{\mathrm{ {ICCV2023}}}$             & ViT-H                 & COCO-LVIS                         & $\textcolor{red}{1.38}$   & $\textcolor{red}{1.50}$          & $1.36$     & $1.75$     & $2.85$    & $4.70$           & $3.41$   & $4.78$  & $1.76$    & $1.98$        \\
\rowcolor[gray]{.9} 	
\textbf{SimpleClick} (ours)          & ViT-H                   & COCO-LVIS                   & $1.40$    & $\textcolor{red}{1.50}$               & $\textcolor{red}{1.33}$      & $\textcolor{red}{1.71}$   & $\textcolor{red}{2.80}$           & $\textcolor{red}{4.65}$   & $\textcolor{red}{3.23}$     & $\textcolor{red}{4.75}$    & $\textcolor{red}{1.72}$  & $\textcolor{red}{1.93}$ \\
\toprule [1pt]
InterFormer~\cite{huang2023interformer}  $_{\mathrm{ {ICCV2023}}}$     & ViT-L                   & COCO-LVIS                  & $1.28$    & $1.36$        & $1.61$     & $2.53$           & $3.25$   & $5.51$          & $4.54$   & $5.21$     & -    & -  \\
\rowcolor[gray]{.9} 	
\textbf{InterFormer} (ours)          & ViT-L                   & COCO-LVIS                   & $\textcolor{red}{1.26}$    & $\textcolor{red}{1.32}$               & $\textcolor{red}{1.56}$      & $\textcolor{red}{2.50}$      & $\textcolor{red}{3.21}$   & $\textcolor{red}{5.46}$       & $\textcolor{red}{4.48}$    & $\textcolor{red}{5.18}$   & - & -\\	
\toprule [1pt]
AdaptiveClick + NFL          & ViT-B                   & COCO-LVIS                   & $\textcolor{red}{1.34}$    & $1.54$               & $1.55$      & $1.92$   & $3.34$           & $5.48$   & $3.66$     & $4.93$    & $2.06$  & $2.39$    \\
\rowcolor[gray]{.9} 	
\textbf{AdaptiveClick} (ours)          & ViT-B                   & COCO-LVIS                   & $\textcolor{red}{1.34}$    & $\textcolor{red}{1.48}$               & $\textcolor{red}{1.40}$      & $\textcolor{red}{1.83}$   & $\textcolor{red}{3.29}$           & $\textcolor{red}{5.40}$   & $\textcolor{red}{3.39}$     & $\textcolor{red}{4.82}$    & $\textcolor{red}{2.03}$ & $\textcolor{red}{2.31}$   \\	
\toprule [2pt]
\end{tabular}}
\label{tab:tab4}
\end{table*}

We also report the segmentation quality achieved with AFL with the COCO-LIVS training dataset.
The results in TABLE~\ref{tab:tab2} showcase that the proposed AFL significantly boosts the IIS model performance when compared to other state-of-the-art loss functions, which is consistent with the outcome observed using the SBD training dataset. Fig.~\ref{fig:fig9} shows the IoU statistics on the test dataset with $4$ clicks. It becomes evident that the proposed AFL always maintains the highest IoU and count values for the same number of hits, which confirms that the proposed AFL has stronger robustness and better accuracy compared with different loss functions.

Overall, the proposed AFL significantly enhances the performance of the IIS model and consistently outperforms existing loss functions designed for IIS tasks.

\textbf{Embedding AFL into State-of-the-Art IIS Methods.}
As shown in TABLE~\ref{tab:tab3}, the proposed AFL brings positive effect improvements for almost all compared methods in all experiments on the test dataset.
This indicates that AFL is robust enough to be applicable to existing mainstream IIS models.
Further, the proposed AFL can bring an average NoC85 and NoC90 performance gain of $0.137$, $0.181$, and $0.119$, $0.103$ for CNN- and transformer-based IIS methods, respectively.
Comparing the performance of FocalClick, SimpleClick, and AdaptiveClick, it can be seen that the proposed AFL is able to deliver NoC85 and NoC90 improvements for FocalClick, SimpleClick, and AdaptiveClick by $0.23$, $0.114$, $0.014$, $0.142$, and $0.098$, $0.07$, respectively, on the five mainstream test datasets.
Comparing the results of CDNet, RITM, and FocusCut, it is clear that the proposed AFL can not only significantly improve the accuracy of the transformer-based model, but also bring performance gains to the CNNs-based model.
This reflects that AFL has more performance gains for transformer-based models and certifies that AFL has the potential to unleash the capacity of vision transformers in IIS tasks.
Consistent with the above analysis, the same results are confirmed in TABLE~\ref{tab:tab4} on the metrics with COCO-LIVS as the training dataset.
In summary, AFL can be applied to different training and test datasets and can adapt to different backbones, loss functions, and different architectures of IIS models with performance gains and strong generalizability.

\subsection{Ablation Studies} \label{sec:exp_ablation}


\subsubsection{Influence of Components in AdaptiveClick}
TABLE~\ref{tab:tab10} and TABLE~\ref{tab:tab9} present the performance changes of the AdaptiveClick with and without the proposed CAMD and AFL.

\textbf{Influence of Components in AdaptiveClick.} As in TABLE~\ref{tab:tab10}, we first show the performance changes with and without the proposed CAMD component on the AdaptiveClick.
The use of the CAMD is effective in improving the performance on the five test datasets.
This positive performance justifies our analysis of the existing mask-fixed model in Sec.~\ref{sec:Pre}.
At the same time, it confirms that the CAMD can effectively solve the interaction ambiguity problem, and the designed CAAM component can provide for long-range propagation between clicks.

\begin{table} [tbp]
\centering
\renewcommand\arraystretch{1.05}
\setlength\tabcolsep{1pt}
\caption{Ablation studies by using or not using CAMD and AFL training on SBD~\cite{majumder2019SBD} and testing on GrabCut~\cite{rother2004grabcut}, Berkeley~\cite{mcguinness2010berkeley}, SBD~\cite{majumder2019SBD}, DAVIS~\cite{perazzi2016DIVAS}, and Pascal VOC~\cite{everingham2010VOC}.}
\resizebox{\linewidth}{!}{\begin{tabular}{cc|cc|cc|cc|cc|cc}
\toprule [2pt]
\multirow{2}{*}{CAMD} & \multirow{2}{*}{AFL}   & \multicolumn{2}{c}{GrabCut} & \multicolumn{2}{c}{Berkeley} & \multicolumn{2}{c}{SBD} & \multicolumn{2}{c}{DAVIS}  & \multicolumn{2}{c}{Pascal VOC} \\	
\cline{4-12}			
& &  NoC85 & NoC90 & NoC85  & NoC90 & NoC85 & NoC90  & NoC85 & NoC90  & NoC85 & NoC90\\
\toprule [1pt]
&   &    $1.46$   & $1.58$               & $1.56$     & $2.46$   & $3.28$           & $5.32$    & $4.20$  & $5.79$ & $2.42$ & $2.85$            \\
$\surd $ &    & $1.40$   & $1.50$               & $\textbf{1.32}$     & $2.24$   & $3.25$   & $5.28$    & $4.03$  & $5.24$  & $2.30$   & $2.75$         \\
& $\surd $   & $1.42$   & $1.60$               & $1.58$     & $2.35$   & $3.29$   & $5.26$    & $4.15$  & $5.55$  & $2.33$   & $2.73$            \\	
\rowcolor[gray]{.9} 
{$\surd $} &  {$\surd $}      & $\textbf{1.38}$    & $\textbf{1.46}$               & $1.38$      & $\textbf{2.18}$   & $\textbf{3.22}$           & $\textbf{5.22}$   & $\textbf{4.00}$     & $\textbf{5.14}$    & $\textbf{2.25}$  & $\textbf{2.66}$   \\
\toprule [2pt]  
\end{tabular}}
\label{tab:tab10}
\end{table}

\begin{table} [tbp]
\centering
\renewcommand\arraystretch{1.18}
\setlength\tabcolsep{1pt}
\caption{Ablation studies by using or not using ADA and AGR of AFL training on SBD~\cite{majumder2019SBD} and testing on GrabCut~\cite{rother2004grabcut}, Berkeley~\cite{mcguinness2010berkeley}, SBD~\cite{majumder2019SBD}, DAVIS~\cite{perazzi2016DIVAS}, and Pascal VOC~\cite{everingham2010VOC}.}
\resizebox{\linewidth}{!}{\begin{tabular}{cc|cc|cc|cc|cc|cc}
\toprule [2pt]
\multirow{2}{*}{ADA} & \multirow{2}{*}{AGR}   & \multicolumn{2}{c}{GrabCut} & \multicolumn{2}{c}{Berkeley} & \multicolumn{2}{c}{SBD} & \multicolumn{2}{c}{DAVIS}  & \multicolumn{2}{c}{Pascal VOC} \\	
\cline{3-12}			
& &  NoC85 & NoC90 & NoC85  & NoC90 & NoC85 & NoC90  & NoC85 & NoC90  & NoC85 & NoC90\\
\toprule [1pt]
&   &    $1.62$   & $1.84$               & $1.69$     & $2.82$   & $4.11$           & $6.51$    & $4.94$  & $6.50$ & $2.81$ & $3.35$            \\
$\surd $ &    & $1.50$    & $1.60$               & $1.64$        & $2.57$   & $3.83$           & $6.03$     & $4.36$ & $5.78$ & $2.63$ & $3.12$          \\
&  $\surd $  & $1.56$    & $1.68$               & $1.54$        & $2.40$   & $3.27$  & $5.23$     & $4.50$ & $5.82$  & $2.32$  & $2.74$             \\
\rowcolor[gray]{.9} 
{$\surd $} & {$\surd $}      & $\textbf{1.38}$    & $\textbf{1.46}$               & $\textbf{1.38}$      & $\textbf{2.18}$   & $\textbf{3.22}$           & $\textbf{5.22}$   & $\textbf{4.00}$     & $\textbf{5.14}$    & $\textbf{2.25}$  & $\textbf{2.66}$   \\
\toprule [2pt]  
\end{tabular}}
\label{tab:tab9}
\end{table}

\begin{table} [tbp!]
\centering
\renewcommand\arraystretch{1.1}
\setlength\tabcolsep{1.0pt}
\caption{The impact of different hyperparameter settings on AdaptiveClick. We trained our AdaptiveClick on SBD~\cite{majumder2019SBD} and tested on GrabCut~\cite{rother2004grabcut}, Berkeley~\cite{mcguinness2010berkeley}, SBD~\cite{majumder2019SBD}, DAVIS~\cite{perazzi2016DIVAS}, and Pascal VOC~\cite{everingham2010VOC} datasets.}
\resizebox{\linewidth}{!}{\begin{tabular}{c|c|c|c|cc|cc|cc|cc|cc}			
\toprule [2pt]
\multirow{2}{*}{$\lambda_{\mathrm{cli}}$ } & \multirow{2}{*}{$\lambda_{\mathrm{afl}}$ } & \multirow{2}{*}{$\lambda_{\mathrm{dice}}$ } & \multirow{2}{*}{$\mathbf{Q}$} & \multicolumn{2}{c}{GrabCut} & \multicolumn{2}{c}{Berkeley} & \multicolumn{2}{c}{SBD} & \multicolumn{2}{c}{DAVIS} & \multicolumn{2}{c}{Pascal VOC} \\
\cline{5-14}
& & & & NoC85 & NoC90  & NoC85 & NoC90 & NoC85 & NoC90   & NoC85 & NoC90 & NoC85 & NoC90 \\
\toprule [1pt]
\rowcolor[gray]{.9} 
${2.0}$                  & $3.0$ & $5.0$  & $5.0$   & $\underline{1.40}$     & $\underline{1.50}$    & $\underline{1.40}$     & $2.39$    & $\underline{3.30}$          & $5.32$    & $\underline{4.08}$    & $\underline{5.35}$  & $\underline{2.31}$   & $\underline{2.76}$  \\

$4.0$                  & $3.0$ & $5.0$  & $5.0$       & $\underline{1.40}$     & $\underline{1.50}$    & $1.43$     & $\underline{2.33}$    & $3.31$          & $\underline{5.31}$    & $4.10$    & $5.37$  & $\underline{2.31}$   & $\underline{2.76}$  \\
\toprule [1pt]
$2.0$                  & $4.0$ & $5.0$  & $5.0$    & $\underline{1.46}$     & $\underline{1.60}$    & $\underline{1.43}$     & $2.29$    & $3.33$          & $5.33$    & $\underline{3.97}$    & $5.38$  & $2.43$   & $2.84$      \\

\rowcolor[gray]{.9} 
${2.0}$                  & $5.0$ & $5.0$  & $5.0$  & $\textbf{1.32}$     & $\textbf{1.42}$    & $1.52$     & $\underline{2.19}$    & $\underline{3.27}$          & $\underline{5.29}$    & $4.08$    & $\underline{5.27}$  & $\underline{2.30}$   & $\underline{2.73}$    \\
\toprule [1pt]

\rowcolor[gray]{.9} 
${2.0}$                  & ${5.0}$ & $5.0$  & $5.0$ & $\underline{1.40}$     & $1.48$    & $\textbf{1.37}$     & $\underline{2.21}$    & $3.30$          & $\underline{5.35}$    & $\textbf{3.90}$    & $\underline{5.29}$  & $\underline{2.31}$   & $\underline{2.73}$     \\

$2.0$                  & $5.0$ & $3.0$  & $5.0$ & $\underline{1.40}$     & $\underline{1.46}$    & $\textbf{1.37}$     & $\underline{2.21}$    & $\underline{3.29}$          & $\underline{5.35}$    & $4.08$    & $5.40$  & $2.44$   & $2.81$    \\

\toprule [1pt]

\rowcolor[gray]{.9} 
$2.0$                  & $5.0$ & $5.0$   & $10.0$   & $\underline{1.38}$    & $\underline{1.46}$               & $\underline{1.38}$      & $\textbf{2.18}$   & $\textbf{3.22}$           & $\textbf{5.22}$   & $\underline{4.00}$     & $\textbf{5.14}$    & $\textbf{2.25}$  & $\textbf{2.66}$    \\

$2.0$                  & $5.0$ & $5.0$  & $20.0$    & $1.48$     & $1.64$    & $1.47$     & $2.28$    & $3.27$          & $5.24$    & $3.99$    & $5.28$  & $2.28$   & $2.73$   \\

$2.0$                  & $5.0$ & $5.0$  & $100.0$   & $1.40$     & $1.48$    & $\textbf{1.37}$     & $2.21$    & $3.29$          & $5.35$    & $4.08$    & $5.40$  & $2.29$   & $2.74$  \\
\toprule [2pt]     
\end{tabular}}     
\label{tab:tab12}
\end{table}

To further explore the effect of AFL on the IIS model, we report the performance of the AdaptiveClick when only AFL is used.
As seen from the TABLEs, the model's performance on GrabCut and the DAVIS dataset is already close to that of the full AdaptiveClick when only AFL is used. This indicates that the proposed AFL has strong robustness. With both CAMD and AFL, the model's performance greatly improved on all five test datasets, which indicates that the proposed CAMD and AFL can effectively address the problems of ``interaction ambiguity'' in existing IIS tasks.

\textbf{Influence of Components in AFL.}
In TABLE~\ref{tab:tab9}, FL does not perform satisfactorily without using any of the proposed components. In contrast, the model's performance improves clearly on all five test datasets after using the proposed ADA.
This indicates that the proposed ADA can help the model adjust the learning strategy adaptively according to differences in sample difficulty distribution.
At the same time, the first hypothesis proposed in Sec.~\ref{sec:sec3.3}, that giving a fixed $\gamma$ to all training samples of a dataset is a suboptimal option that may prevent achieving satisfactory performance on low-confidence easy pixels, is verified, which proves the rationality and effectiveness of ADA.

Also, we explore the case of using only the AGR component.
The use of AGR results in more significant performance gains on the Berkeley, SBD, and PasvalVOC test datasets compared to the ADA-only.
This indicates the effectiveness of the proposed AGR and validates the second hypothesis proposed in Sec.~\ref{sec:sec3.3}, that FL has the problem that when one wants to increase the concentration on learning with severe hard-easy imbalance, it tends to sacrifice part of low-confidence easy pixels’ loss contribution in the overall training process. At the same time, this proves the validity of the theoretical analysis in Sec.~\ref{sec:sec3.3}.
Namely, the gradient direction of the simulated BCE can help AFL classify low-confidence easy pixels.

Finally, when both ADA and AGR are used, AFL brings a minimum gain of $0.14$, $0.34$ (on GrabCut), and a maximum gain of $0.94$, $1.23$ (on DAVIS and SBD) over FL on the NoC85 and NoC90, respectively. Such significant performance improvements again demonstrate the effectiveness of the AFL.

\subsubsection{Influence of Hyper-parameters} \label{Hyper-parameters}

In TABLE~\ref{tab:tab12} and TABLE~\ref{tab:tab13}, the effects of $\lambda_{\mathrm{cli}}$, $\lambda_{\mathrm{afl}}$, $\lambda_{\mathrm{dice}}$, $\mathbf{Q}$, and $\lambda_{\mathrm{mask}}$ on the model performance are explored.
Then, TABLE~\ref{tab:tab11} and TABLE~\ref{tab:tab14} report the results of different $\delta$ and $\alpha$ weights.

\begin{table} [tbp!]
\centering
\renewcommand\arraystretch{1.05}
\setlength\tabcolsep{2pt}
\caption{The performance change by different weights of $\lambda_{\mathrm{mask}}$ trained on SBD~\cite{majumder2019SBD} and tested on GrabCut~\cite{rother2004grabcut}, Berkeley~\cite{mcguinness2010berkeley}, SBD~\cite{majumder2019SBD}, DAVIS~\cite{perazzi2016DIVAS}, and Pascal VOC~\cite{everingham2010VOC} datasets.}
\resizebox{\linewidth}{!}{\begin{tabular}{c|cc|cc|cc|cc|cc}			
\toprule [2pt]
\multirow{2}{*}{$ {\lambda_{\mathrm{mask}}}$} & \multicolumn{2}{c}{ {GrabCut}} & \multicolumn{2}{c}{ {Berkeley}} & \multicolumn{2}{c}{ {SBD}} & \multicolumn{2}{c}{ {DAVIS}} & \multicolumn{2}{c}{ {Pascal VOC}} \\
\cline{2-11}
&  {NoC85} &  {NoC90}  &  {NoC85} &  {NoC90} &  {NoC85} &  {NoC90}   &  {NoC85} &  {NoC90} &  {NoC85} &  {NoC90} \\
\toprule [1pt]
$ {0.5}$  & $ {1.42}$    & $ {1.58}$    & $ {1.45}$      & $ {2.27}$   & $ {3.33}$           & $ {5.32}$   & $ {\textbf{3.81}}$     & $ {5.30}$    & $ {2.35}$  & $ {2.76}$    \\

\rowcolor[gray]{.9} 
$ {1.0}$  & $ {\textbf{1.38}}$    & $ {\textbf{1.46}}$               & $ {\textbf{1.38}}$      & $ {\textbf{2.18}}$   & $ {\textbf{3.22}}$           & $ {\textbf{5.22}}$   & $ {4.00}$     & $ {\textbf{5.14}}$    & $ {\textbf{2.25}}$  & $ {\textbf{2.66}}$ \\

$ {1.5}$     & $ {1.44}$    & $ {1.56}$    & $ {1.57}$      & $ {2.47}$   & $ {3.34}$           & $ {5.36}$   & $ {4.21}$     & $ {5.41}$    & $ {2.30}$  & $ {2.74}$     \\

$ {2.0}$  & $ {1.52}$    & $ {1.66}$    & $ {1.60}$      & $ {2.54}$ & $ {3.43}$     & $ {5.46}$  & $ {4.31}$           & $ {5.60}$       & $ {2.34}$  & $ {2.81}$\\
\toprule [2pt]     
\end{tabular}}
\label{tab:tab13}
\end{table}

\begin{table} [tbp!]
\centering
\renewcommand\arraystretch{1.0}
\setlength\tabcolsep{2pt}
\caption{The performance change by using AFL under different weights of $\delta$ trained on SBD~\cite{majumder2019SBD} and tested on GrabCut~\cite{rother2004grabcut}, Berkeley~\cite{mcguinness2010berkeley}, SBD~\cite{majumder2019SBD}, DAVIS~\cite{perazzi2016DIVAS}, and Pascal VOC~\cite{everingham2010VOC}.}
\resizebox{\linewidth}{!}{\begin{tabular}{c|cc|cc|cc|cc|cc}			
\toprule [2pt]
\multirow{2}{*}{$\delta $ } & \multicolumn{2}{c}{GrabCut} & \multicolumn{2}{c}{Berkeley} & \multicolumn{2}{c}{SBD} & \multicolumn{2}{c}{DAVIS} & \multicolumn{2}{c}{Pascal VOC} \\
\cline{2-11}
& NoC85 & NoC90  & NoC85 & NoC90 & NoC85 & NoC90   & NoC85 & NoC90 & NoC85 & NoC90 \\
\toprule [1pt]
$0.0$  & $\textbf{1.36}$    & $\textbf{1.46}$    & $1.41$      & $2.31$   & $3.21$           & $5.18$   & $4.01$     & $5.20$    & $2.26$  & $2.69$  \\

$0.1$                      & $1.46$     & $1.56$              & $1.50$     & $2.33$    & $3.26$          & $5.26$    & $4.18$   & $5.29$  & $2.32$   & $2.74$     \\

$0.2$                       & $1.38$     & $1.48$               & $1.45$      & $2.31$      & $3.24$          & $5.18$ & $4.17$     & $5.30$  & $2.32$  &$2.74$          \\

\rowcolor[gray]{.9} 
$0.4$     & $1.38$    & $\textbf{1.46}$               & $\textbf{1.38}$      & $\textbf{2.18}$   & $3.22$           & $5.22$   & $\textbf{4.00}$     & $\textbf{5.14}$    & $\textbf{2.25}$  & $\textbf{2.66}$     \\

$0.6$                       & $1.42$     & $1.60$     & $1.58$      & $2.35$     & $\textbf{3.19}$	 & $\textbf{5.16}$	 & $4.15$	 & $5.25 $	 & $2.30$	& $2.73$   \\
\toprule [2pt]     
\end{tabular}}     
\label{tab:tab11}
\end{table}

\textbf{Hyper-parameter Analysis of AdaptiveClick.}
In TABLE~\ref{tab:tab12}, we validate the performance of our model on the five test datasets when $\lambda_{\mathrm{cli}}$ is set to $2$ and $4$ respectively. We observe that the model is not sensitive to the value of $\lambda_{\mathrm{cli}}$, but the model demonstrates better results when $\lambda_{\mathrm{cli}}$ is $2$.
Consequently, in this study, we select a click weight value of $\lambda_{\mathrm{cli}} = 2$.
Furthermore, when comparing the performance for varying AFL weight values of $\lambda_{\mathrm{afl}} = \{5, 4, 3\}$, we observe significant fluctuations in the model's performance as $\lambda_{\mathrm{afl}}$ decreases. 
As a result, we opt for an AFL weight of $\lambda_{\mathrm{afl}} = 5$.
Then, observing the performance change when the dice weight of $\lambda_{\mathrm{dice}}$ is taken as $5$, $4$, and $3$, we can see that the model performs most consistently when $\lambda_{\mathrm{dice}}$ is $5$.
Further, we tested the effect of the number of clicks $\mathbf{Q}$ on the new performance of the model, and the number of $\mathbf{Q}$ chosen was $5$, $10$, $20$, and $100$, respectively.
The experimental results show that the model has optimal performance when $\mathbf{Q}=10$. 
Finally, the results of different mask loss weights, $\lambda_{\mathrm{mask}}$, are listed in TABLE~\ref{tab:tab13}. When $\lambda_{\mathrm{mask}}=1.0$, AdaptiveClick has the best performance. 
As a result, the values are set to $\lambda_{\mathrm{mask}}=1.0$, $\lambda_{\mathrm{cli}}=2.0$, $\lambda_{\mathrm{afl}}=5.0$, $\lambda_{\mathrm{dice}}=5.0$, and $\mathbf{Q}=10.0$.

\begin{table} [tbp!]
\centering
\renewcommand\arraystretch{1.05}
\setlength\tabcolsep{2pt}
\caption{The performance change by using AFL under different weights of $\alpha$ trained on SBD~\cite{majumder2019SBD} and tested on GrabCut~\cite{rother2004grabcut}, Berkeley~\cite{mcguinness2010berkeley}, SBD~\cite{majumder2019SBD}, DAVIS~\cite{perazzi2016DIVAS}, and Pascal VOC~\cite{everingham2010VOC}.}
\resizebox{\linewidth}{!}{\begin{tabular}{c|cc|cc|cc|cc|cc}			
\toprule [2pt]
\multirow{2}{*}{$ {\alpha}$} & \multicolumn{2}{c}{ {GrabCut}} & \multicolumn{2}{c}{ {Berkeley}} & \multicolumn{2}{c}{ {SBD}} & \multicolumn{2}{c}{ {DAVIS}} & \multicolumn{2}{c}{ {Pascal VOC}} \\
\cline{2-11}
&  {NoC85} &  {NoC90}  &  {NoC85} &  {NoC90} &  {NoC85} &  {NoC90}   &  {NoC85} &  {NoC90} &  {NoC85} &  {NoC90} \\
\toprule [1pt]
$ {0.0}$  & $ {1.44}$    & $ {1.56}$    & $ {1.53}$      & $ {\textbf{2.15}}$   & $ {3.43}$           & $ {5.47}$   & $ {\textbf{3.93}}$     & $ {5.28}$    & $ {2.32}$  & $ {2.76}$    \\

$ {0.5}$  & $ {1.46}$    & $ {1.60}$    & $ {1.43}$      & $ {2.30}$   & $ {3.29}$           & $ {5.29}$   & $ {3.99}$     & $ {5.27}$    & $ {\textbf{2.23}}$  & $ {\textbf{2.66}}$    \\

\rowcolor[gray]{.9} 
$ {1.0}$  & $ {\textbf{1.38}}$    & $ {\textbf{1.46}}$               & $ {\textbf{1.38}}$      & $ {2.18}$   & $ {\textbf{3.22}}$           & $ {\textbf{5.22}}$   & $ {4.00}$     & $ {\textbf{5.14}}$    & $ {2.25}$  & $ {\textbf{2.66}}$\\

$ {1.5}$     & $ {\textbf{1.38}}$    & $ {1.48}$    & $ {1.61}$      & $ {2.53}$   & $ {3.26}$           & $ {5.30}$   & $ {4.11}$     & $ {5.45}$    & $ {2.26}$  & $ {2.67}$     \\

$ {2.0}$  & $ {1.40}$    & $ {1.48}$    & $ {1.57}$      & $ {2.48}$   & $ {3.29}$           & $ {5.32}$   & $ {4.06}$     & $ {5.38}$    & $ {2.30}$  & $ {2.76}$\\
\toprule [2pt]     
\end{tabular}}
\label{tab:tab14}
\end{table}

\begin{figure}[t]
\centering
\includegraphics[width=0.45 \textwidth]{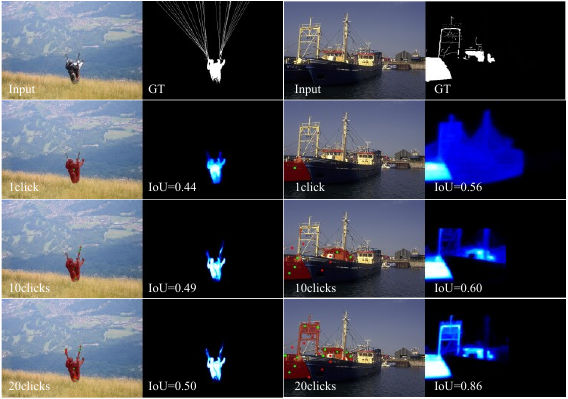}

\caption{Two illustrations of failure cases.}
\label{fig:failure}
\end{figure}

\textbf{Hyper-parameter analysis of AFL.} As shown in TABLE~\ref{tab:tab11} and TABLE~\ref{tab:tab14}, when $\delta$ is set to $0$ and $\alpha$ is $0$, AFL can be approximated as the value of NFL after applying ADA. Under these conditions, the NoC85 and NoC90 results of AFL are suboptimal. However, as $\delta$ increases from $0$ to $0.2$ and $\alpha$ rises from $0$ to $0.5$, there is a clear improvement in performance across all five datasets, providing compelling evidence for the efficacy of AFL. 
When $\delta=0.4$ and $\alpha=1.0$, the values of evaluation metrics start to fluctuate. 
Consequently, based on the outcomes of our experiments, the settings of $\delta = 0.4$ and $\alpha = 1.0$ emerge as the preferred parameters.

\section{Discussion} \label{sec:discussion}
We introduce AdaptiveClick to effectively address the challenges associated with ``interaction ambiguity'', focusing on both inter-class click ambiguity resolution and intra-class click ambiguity optimization. Benefiting from CAMD's sensitivity to clicks and AFL's reduction of ``gradient swamping'', AdaptiveClick exhibits a more competitive performance on IIS across nine datasets. In addition to the IIS task, AdaptiveClick offers the following two advantages: 1) AdaptiveClick can offer a robust baseline for IIS part from the generalized prompt segmentation models~\cite{kirillov2023segany, zou2023segment}, fostering the advancement of this task; 2) whether referring~\cite{BRINet,chen2023epcformer}, interactive~\cite{liu2022simpleclick,RanaMahadevan23DynaMITe}, or prompt~\cite{kirillov2023segany, zou2023segment} image/video segmentation tasks, ``gradient swamping'' is often present during the loss optimization process. Our validation of AFL's effectiveness and generality for IIS tasks supports the notion that it can provide potential optimization guarantees in various fields.

However, AdaptiveClick has two limitations: 1) As depicted in Fig.~\ref{fig:failure}, AdaptiveClick may not be effective for slender objects with heavy occlusions, leading to potential segmentation failures; and 2) Similar to other transformer-based IIS methods, our approach may not be efficient for low-power devices. Consequently, enhancing the segmentation of slender objects with heavy occlusion through refined mask or pipeline optimization strategies and exploring lightweight transformer-based IIS frameworks through model compression and knowledge distillation represent two promising avenues. We defer these potential improvements to future work.

\section{Conclusion} \label{sec:conclusion}
In this paper, we rethink the ``interaction ambiguity'' problem in the IIS task from the perspective of inter-class click ambiguity resolution to intra-class click ambiguity optimization. First, we designed CAMD to enhance the long-range interaction of clicks in the forward pass process and introduce a mask-adaptive strategy to find the optimal mask matching with clicks, thus solving the inter-class interaction ambiguity. Second, we observe that the root of the intra-class hard pixel misclassification is ``gradient swamping'' and propose a new AFL based on the gradient theory of BCE and FL to enforce the network to pay more attention to the ambiguous pixels, reducing the intra-class click ambiguity. Finally, experiments on nine datasets of the IIS task show that the proposed AdaptiveClick yields state-of-the-art performances.

{\small
\bibliographystyle{IEEEtran}
\bibliography{refs}
}

\clearpage
\appendices

\section{Adaptive Focal Loss}
The proposed Adaptive Focal Loss (AFL) possesses the following qualities:
 
\begin{figure}[htbp]
    \centering
    \subfigure[AFL with different $\gamma$.]{
    \includegraphics[width=0.244\textwidth]{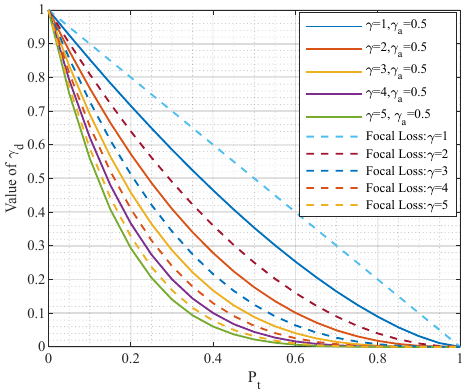} \label{fig:fig5a}}\subfigure[AFL with different $\gamma_{\mathrm{a}}$.]{
    \includegraphics[width=0.229\textwidth]{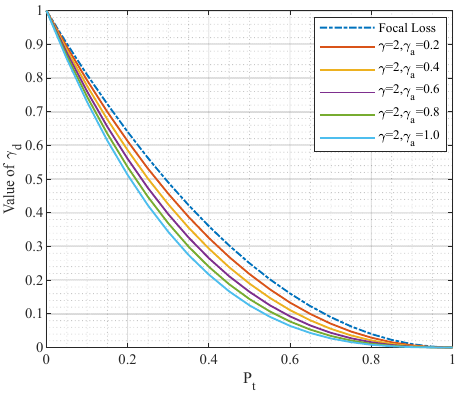}\label{fig:fig5b}}\caption{The variation of $\gamma_{\mathrm{d}}$ at different values of the proposed AFL. In Fig. \ref{fig:fig5a}, to explore the effect of $\gamma$ on AFL, where $\gamma_{a}$ = 0.5. Similarly, in Fig. \ref{fig:fig5b}, $\gamma$ is set to $2$ to explore the effect of different values of $\gamma_{\mathrm{a}}$ on AFL.
    \label{fig:fig5}}
\end{figure}

\textsc{(1) AFL allows the IIS model to focus more on low-confidence easy samples.} As in Fig.~\ref{fig:fig5a}, the AFL allows pixels with smaller $\mathbf{P}^{i}_{\mathrm{t}}$ to obtain higher gradient values for the same value of $\gamma$ compared with Focal Loss~\cite{lin2017focal};

\textsc{(2) AFL allows models to adaptively switch learning strategies.} As in Fig.~\ref{fig:fig5b}, compared to the rigid learning strategy of Focal Loss~\cite{lin2017focal}, AFL is more flexible in its learning style. When faced with samples with significant differences in pixel difficulty distributions, the value of $\gamma_{\mathrm{d}}$ will be increased to overcome inadequate learning. Conversely, when facing samples with minor differences, the proposed $\gamma_{\mathrm{d}}$ can adopt a small value to overcome over-learning (over-fitting) of extremely difficult pixels.

\end{document}